\documentclass[journal]{IEEEtran}

\usepackage[pdftex]{graphicx}

\usepackage{amssymb}
\usepackage{latexsym}
\usepackage{cite}
\usepackage[hidelinks]{hyperref}
\usepackage{bm}
\usepackage{todonotes}
\usepackage{gensymb} 
\usepackage{amsmath} 
\usepackage{caption} 
\usepackage{subfigure}
\usepackage{float}
\usepackage{cuted}
\usepackage{graphbox}

\newcommand{\comment}[1] {}
\newcommand{\comments}[1] {}

\newlength \figwidth
\if@twocolumn
\else
\fi
\def\figratio1{.1}

\begin{document}

\title{SymNet: Symmetrical Filters in Convolutional Neural Networks}

\author{Gregory Dzhezyan~\IEEEmembership{Member,~IEEE} and Hubert Cecotti~\IEEEmembership{Senior Member,~IEEE}
\thanks{G. Dzhezyan H. Cecotti are with the Department of Computer Science, College of Science and Mathematics, Fresno State University, Fresno, Ca, USA.}
\thanks{Manuscript received December X, 201X; revised August XX, 20XX. 
Corresponding author: H. Cecotti (email: hcecotti@csufresno.edu).}} 

\maketitle


\begin{abstract}
Symmetry is present in nature and science. In image processing, kernels for spatial filtering possess some symmetry (e.g. Sobel operators, Gaussian, Laplacian). Convolutional layers in artificial feed-forward neural networks have typically considered the kernel weights without any constraint. In this paper, we propose to investigate the impact of a symmetry constraint in convolutional layers for image classification tasks, taking our inspiration from the processes involved in the primary visual cortex and common image processing techniques. The goal is to assess the extent to which it is possible to enforce symmetrical constraints on the filters throughout the training process of a convolutional neural network (CNN) by modifying the weight update preformed during the backpropagation algorithm and to evaluate the change in performance. The main hypothesis of this paper is that the symmetrical constraint reduces the number of free parameters in the network, and it is able to achieve near identical performance to the modern methodology of training. In particular, we address the following cases: x/y-axis symmetry, point reflection, and anti-point reflection. The performance has been evaluated on four databases of images. The results support the conclusion that while random weights offer more freedom to the model, the symmetry constraint provides a similar level of performance while decreasing substantially the number of free parameters in the model. Such an approach can be valuable in phase-sensitive applications that require a linear phase property throughout the feature extraction process.
\end{abstract}


\IEEEpeerreviewmaketitle

\section{Introduction}
\label{sec:introduction}
		
Convolutional Neural Networks (CNNs) are a class of Artificial Neural Networks (ANNs) commonly used in deep learning~\cite{goodfellow2017}. Their prevalence today in computer vision tasks is unprecedented, and rightfully so, as they have demonstrated extraordinary ability in challenging pattern recognitions tasks, most notably for object recognition \cite{gonzalez2008,russell2009}. In addition, recent results suggest that it is better to learn everything, with a shift from scale-invariant feature transform (SIFT)~\cite{lowe1999} based features to CNN based methods for instance retrieval applications. The trend is towards end-to-end feature learning and extraction approaches~\cite{zheng2018}. While it has been shown that it is better to learn features, there is key evidence that pre-training helps with deep learning~\cite{erhan2010}. In addition, the optimal choice of architecture including the number of layers, feature maps, and convolutional layer settings, remain a challenge given the large number of architectures that have been proposed in recent years.
Today, researchers from a variety of backgrounds, including neural engineering~\cite{cecotti2011}, robotics~\cite{redmon2015}, chemistry~\cite{ma2015}, and astronomy~\cite{dieleman2015}, are employing CNNs to not only advance understanding within their own fields but to produce trained networks that may viable for use commercially. This is evident in society as manufacturers of devices, such as for Internet of Things (IoT) applications, have begun to include highly specialized integrated circuits that are well adapted for CNN execution~\cite{hagiwara2018}. As we move forward with the deployment of CNNs into embedded applications, it is necessary to examine the efficiency of CNNs in terms of number of free parameters, as this has direct physical implications for embedded systems.

CNNs are typically known to require a large number of labeled examples to capture the variability that exists across examples, in particular to provide features that are tolerant to different types of transformations, e.g. translation and rotation. For computer vision tasks, the architectures of CNNs are typically based on the principle of the human primary visual cortex (V1)~\cite{grillspector2004}, where the goal is to preprocess the input image to extract edges or to enhance some particular features. For instance, before CNNs demonstrated their superiority for feature extraction through transfer learning, it was common to consider Gabor filters to preprocess images by selecting a set of frequencies and orientations~\cite{jain1997,kamarainen2006,shen2009}. Such an approach places a significant bias on the features due to the arbitrary choice of the frequencies and orientations. 

This paper investigates parameter reduction in CNNs by attempting to enforce a symmetrical constraint on the learned weights within convolutional layers of a network. Symmetry is ever present in many natural and scientific phenomena, including modern physics~\cite{zee2007}, and from the perspective of information representation, symmetry has the potential to reduce complexity by compacting it into lighter structures. Moreover, the rationale of this paper is rooted from the fact that many state of the art spatial filtering techniques, for edge detection (e.g. Sobel, Prewitt, Gabor), smoothing (e.g. Gaussian filter), and image enhancement (e.g. Mexican Hat, Difference of Gaussian, Laplacian of Gaussian), have symmetrical properties that can be taken into account through training convolutional layers. For instance, the derivatives at high frequencies are useful for edge detection~\cite{basu2002,ziou1998}. Since edges are known to be meaningful features, it is plausible that a CNN may eventually approximate a symmetrical filter in order to learn to classify the images it gets as an input. To enforce a symmetrical constraint throughout the learning process, a change to the backpropagation algorithm is required so that it adheres to specific weight update algorithms implemented for several filters, each with different forms of symmetry. First, the proposed filters are first initialized in a way that the symmetrical properties are set. Second, the back propagated errors are combined to satisfy the constraints of the filter kernels.  

A finite impulse response (FIR) filter is linear-phase if and only if its coefficients are symmetrical around the center coefficient. Then, symmetric filters provide a linear phase, corresponding to the condition where the phase response of the filter is a linear function of frequency. With such a filter, all frequency components of the input image are shifted in time by the same constant amount (the group delay). There is therefore no phase distortion due to the time delay of frequencies relative to one another~\cite{roberts2011} ~\cite{sundararajan2003}. A filter will delay different frequency components of a signal by the same amount if the filter has linear phase. Such a property in filters can be desirable as applications without linear phase can introduce artifacts in images.

The main contributions of this paper are:
\begin{itemize}
\item Three types of symmetrical 2D filters that can constraint the convolutional layers to extract specific types of features. These types of filters are linear phase and correspond to Type I (Even-order symmetric coefficients) and Type III Even-order (antisymmetric coefficients) FIR filters.
\item A new way to reduce the number of free parameters in CNNs through weight sharing of the weights within a filter, which decreases the computational cost of the forward operation.
\end{itemize}	

The remainder of the paper is organized as follows. First, we begin by discussing relevant works in Section~\ref{sec:related-works}, then give brief descriptions of the forward and backward propagation procedures for CNNs in Section~\ref{sec:methods} in order to facilitate a clear foundation for our description of symmetric filters that follows. Descriptions of the databases and the network architectures used for testing are covered in Sections~\ref{sec:database-descriptions} and ~\ref{sec:architecture}. The results are presented in Section~\ref{sec:results} and their impact on CNNs are discussed in Section~\ref{sec:discussion}.
		
\section{Related Works}
\label{sec:related-works}

Today's state of the art convolutional neural networks are heavily influenced by the work of LeCun~\cite{lecun1990} and the neocognitron~\cite{fukushima1980}. In 1998, LeCun proposed the LeNet-5 architecture, which was able to successfully classify images from a large dataset of handwritten digits~\cite{lecun1998}. This architecture combined convolutional layers, followed by pooling layers, and then terminated the network with fully connected layers. Most notably LeCun introduced the use of the backpropagation algorithm to ConvNets which allowed for the modification of the weights for the entire network based on the error calculated at the output. While LeNet used average pooling in its architecture, it has been shown that average pooling was not as robust as max pooling in their HMAX model~\cite{riesenhuber1999}. The argument being made was that average pooling would actually obscure features because the responses of simple cells were being summed, while the max operation simply returned strongest response and therefore had the best chance of detecting features. Additionally, average pooling was shown to be variant to scale, due to the response strength after pooling being correlated with object size.

One such influential ConvNet architecture inspired by LeNet was the AlexNet architecture that received critical acclaim due to breaking the record at the ImageNet Large Scale Visual Recognition Challenge (ILSVRC)~\cite{krizhevsky2012}. In addition to max pooling layers, AlexNet changed the activation function used by neurons in each layer. Previously, it was common to use a saturating non-linearity as the activation function, such as the hyperbolic tangent or sigmoid, however AlexNet used ReLU or Rectified Linear Unit as its activation function as this function provides a significant performance boost to the training of the networks. They claimed a six times speed improvement to reach the same error rate with a network using ReLU compared to one using the hyperbolic tangent. AlexNet reduced overfitting by utilizing dropout layers, where neurons in some given layer are randomly selected to have their weights zeroed, effectively changing the number of neurons in that layer~\cite{srivastava2014}, and augmenting the data set by translating and reflecting images in the training set. 

In the wake of AlexNet came the discovery that building networks with a larger number of layers increased the performance of the network. The performance on Very Deep Convolutional networks showed that improvements could be achieved by using small convolutions (3 $\times$ 3) and increasing the network size to encompass 16-19 layers~\cite{simonyan2014}. However, increasing the number layers increases the number of parameters required to learn. GoogLeNet was an even deeper network utilizing 22 layers~\cite{szegedy2014}. Yet, it includes the inception module in which multiple convolutions of different filter sizes are employed: 1 $\times$ 1, 3 $\times$ 3, and 5 $\times$ 5 convolutions are computed in parallel. Additionally, a pooling operation is conducted within the module, in parallel with the other operations, for good measure. The output of all the convolutions are then concatenated together along the depth dimension. However, preforming these extra convolutions in parallel did not end up increasing the number parameters to learn due to the fact that Google introduced a 1 $\times$ 1 convolution to be preformed on the input to another larger convolution to reduce the dimensionality of inputs. These works stress the importance of the architecture and the parameters related to the convolutional layers.
	
\section{Methods}
\label{sec:methods}

\subsection{CNN Forward Propagation}

Forward propagation for a convolutional layer in a CNN involves either performing a 2D convolution or cross-correlation on an image inputted to the layer. Convolution and cross-correlation are similar processes and from the perspective of the artificial neural network are indistinguishable. Both operations involve taking a small matrix of weights (kernel/filter) of size $N_w \times N_w$, overlaying them with section of an image and summing the element wise multiplication of the weights and image intensities directly under the filter. The filter is typically translated across the entire image such that the operation has been executed at nearly every pixel. In order for the operation to be considered a convolution, its required that the filter be rotated by $180 \degree$ before applying it to the image, otherwise it is a cross-correlation, expressed as follows: 
\begin{eqnarray}
 y(i,j) & = &  \sum\limits_{\substack{i1=-N_{half},\\j1=-N_{half}}}^{\substack{N_{half},\\N_{half}}} w_{i1,j1} \cdot x(i+i1,j+j1)
\end{eqnarray}
where $N_{half}=\left\lfloor N_w/2 \right\rfloor$.

After a convolution is performed on an image of size $N_{in} \times N_{in}$ with a filter of size $N_w \times N_w$, the resultant image shape can be computed with Eq.~\ref{equ:conv-dim}.
\begin{eqnarray}
N_{out} & = &	\frac{N_{in} + 2N_P - N_w}{N_S} + 1
	\label{equ:conv-dim}
\end{eqnarray}
where $N_P$ indicates how much padding is added to the image before convolution and $N_S$ is the stride taken by the filter~\cite{gonzalez2008}. Convolution and cross-correlation can be executed forming Toeplitz matrices from the filters and unrolling the image into a column vector. The filter is transformed into a Toeplitz matrix by inflating with zeros until it becomes the same shape as the input image. Then this inflated filter is unrolled into a row vector. Shifted versions of the filter vector are then copied in as rows of a Toeplitz matrix. A matrix product between the Toeplitz representation of the filter and the unrolled image yields the result of the cross-correlation operation. Using the example image $\mathbf{I}$ and filter $\mathbf{W}$, we can form a Toeplitz matrix from $\mathbf{W}$ by first expanding it to the shape of $\mathbf{I}$ and filling with zeros. Then, unroll $ \mathbf{W_{expanded}}$ and copy shifted versions of it into a new matrix, which will be the Toeplitz matrix. The shape of the Toeplitz matrix for an image of shape $ N_1 \times N_2$ and filter of size $N_w \times N_w$ can be determined by Eq.~\ref{equ:toe-dim}
\begin{equation}
	\left( \frac{N_1 + 2P - N}{S} + 1 \right)\left( \frac{N_2 + 2P - N}{S} + 1 \right) \times \left(N_1N_2\right)
	\label{equ:toe-dim}
\end{equation}
The result $\mathbf{R}$ of the convolution is computed as follows:
\begin{equation}
\mathbf{R} = \mathbf{W_{toe}} \mathbf{I} 
\end{equation}
where $\mathbf{I}$ is a column vector of the unrolled image and $\mathbf{R}$ is a column vector, which needs to be reshaped back into a 2D matrix. Finally, the forward propagation equation for convolutional layer is:
\begin{equation}
\mathbf{o}^{l} = f_l \left(   \mathbf{W_{toe}}^{l} \mathbf{o}^{l-1} \right)	
\label{equ:cnn-forward}
\end{equation}
where $\mathbf{o}^{l}$ is the output for layer $l$, $f_l(\cdot)$ is the activation function for layer $l$, $ \mathbf{W_{toe}}^{l}$ is the filter Toeplitz matrix (weight matrix) that connects layer $l-1$ to layer $l$, and $\mathbf{o}^{l-1}$ is the column vector of the output from layer $l-1$.

\subsection{CNN Backpropagation}

Backpropagation for a convolutional layer in a CNN is separated out into two steps: the error backpropagation and the weight update. In the error backpropagation step, we compute the error at a previous layer using the error at the current layer. For the computed error at a layer of the network:
\begin{equation}
	\mathbf{e}^l = \begin{bmatrix} e^l_0 & e^l_1 & \cdots & e^l_m \end{bmatrix}
\end{equation}
We can back propagate the error by multiplying $\mathbf{e}^l$ by the filter Toeplitz matrix $\mathbf{ W_{toe} }^{l}$ that connects layer $l-1$ to layer $l$ and then performing a dot product with the derivative of the activated outputs of layer $l-1$.
\begin{equation}
	\mathbf{e}^{l-1} = \left( \mathbf{e}^l  \mathbf{W_{toe}}^{l} \right) \cdot f'_{l-1}(\mathbf{o}^{l-1})
	\label{equ:cnn-err-backprop}
\end{equation}
where $f'_{l-1}(\cdot)$ is the derivative of the activation function for layer $l-1$.

In order to perform the weight update, we take the error for current layer $\mathbf{e}^l$, reshape it into the same shape as the output of layer $l$, and perform a cross-correlation with the input to the layer (i.e the output of the previous layer $\mathbf{o}^{l-1}$), which is reshaped into a 2D matrix.

\begin{equation}
	\Delta \mathbf{ W } = \alpha \left( \mathbf{E}^l \star \mathbf{ O }^{l-1}  \right)
	\label{equ:cnn-weight-update}
\end{equation}
where $\alpha$ is the learning rate, $\mathbf{E}^l$ is the error vector $\mathbf{e}^l$ reshaped to be the same shape as the output for layer $l$, and $\mathbf{ O }^{l-1}$ is the output vector of the previous layer $\mathbf{o}^{l-1}$ reshaped into the same shape as the output shape of layer $l-1$. It is worth noting that if a stride is greater than 1 ($S > 1$) was used in the forward convolution, then this must be accounted for in Eq.~\ref{equ:cnn-weight-update} by appropriately extending $\mathbf{E}^l$ with $S-1$ columns and rows of zeros inserted between each of its column and row.

\subsection{Linear phase FIR filters}

FIR filters are filters with a finite duration. An $r^{th}$ order discrete FIR filter lasts $k+1$ time points; the number of taps is the same as the filter length. We denote by $N_b$ the size of the filter ($N_b=k+1$). The discrete convolution is expressed by:
\begin{eqnarray}
y(n) & = & \sum\limits_{i=0}^{r} b_i \cdot x(n-i)
\end{eqnarray}	
where $x$ and $y$ are the input and output signals, respectively; $k$ is the filter order, $b_i$ represents the weight of the filter at time $i$, $0 \leq i \leq r$. 

Linear phase FIR filters are divided into four types: Type I (even-order, symmetric coefficients), Type II (odd-order, symmetric coefficients), Type III (even-order antisymmetric coefficients), and Type IV (odd-order antisymmetric coefficients). Types III and IV can be used to design differentiators~\cite{oppenheim1989}, which can be used for edge detection.
The symmetry of the impulse response is written as: $w_n = w(N_w-1-n)$ (Type I and II), and the anti-symmetry is written as: $b_n = -b(N_b-1-n)$ (Type III and IV). The parameters of the filter correspond to an even function centered on N/2 for Type I and II, and an odd function for Type III and IV.
In this paper, we will focus on Type I and III as they correspond to kernel sizes that are typically used in the literature for setting the input windows of convolutional layers, e.g. 3 $\times$ 3, 5 $\times$ 5.
We denote by $A(\omega)$ and $\theta(\omega)$ the amplitude response and the continuous phase function of the filter, respectively. 
A linear phase filter is defined by its frequency response:
\begin{eqnarray}
H^f(\omega) & = & A(\omega) \cdot e^{j\theta(\omega)}
\end{eqnarray}
with 
\begin{eqnarray}
\theta(\omega)& = & -M\omega + B
\end{eqnarray}
where $j$ is the imaginary unit.

FIR filters Type I of length $N_w$ are defined as follows: 
\begin{eqnarray}
A(\omega) & = & h(M) + 2\sum\limits_{n=0}^{M-1} w_n \cdot cos((M-n)\omega) \\
\theta(\omega) & = & -M\omega  \nonumber 
\end{eqnarray}
where $M=(N-1)/2$.
For a filter of length 5, the filter can be expressed by:
\begin{eqnarray}
H^{f} & = &  b_0 + b_1e^{-j\omega} + b_2 e^{-2j\omega} + w_1 e^{-3j\omega} + b_0 e^{-4\omega} \\
    & = & e^{-2j\omega} (2b_0 \cdot cos(2\omega)+ 2b_1 \cdot cos(\omega)+b_2) \nonumber \\
    & = & A(\theta)e^{j\theta(\omega)} \nonumber
\end{eqnarray}
with $\theta(\omega)=-2\omega$ and $A(\omega)=2b_0 \cdot cos(2\omega)+ 2b_1 \cdot cos(\omega)+b_2$.

FIR filters Type III of length $N$ are defined as follows: 
\begin{eqnarray}
A(\omega) & = & 2\sum\limits_{n=0}^{M-1} b_n \cdot sin((M-n)\omega) \\
\theta(\omega) & = & -M\omega + \pi/2  \nonumber 
\end{eqnarray}
where $M=(N-1)/2$.
For a filter of length 5, the filter can be expressed by:
\begin{eqnarray}
H^{f} & = &  b_0 + b_1e^{-j\omega} - b_1 e^{-3j\omega} - b_0 e^{-4\omega} \\
    & = & e^{-2j\omega} e^{j\pi/2}  (2b_0 \cdot sin(2\omega)+2b_1 \cdot sin(\omega)) \nonumber \\
    & = & A(\theta)e^{j\theta(\omega)} \nonumber
\end{eqnarray}
with $\theta(\omega)=-2\omega+\pi/2$ and $A(\omega)=2b_0 \cdot sin(2\omega)+2b_1 \cdot sin(\omega)$.

\subsection{Symmetric Receptive Fields}

\begin{figure}[h!]
	\centering	    
	\includegraphics[width=8cm]{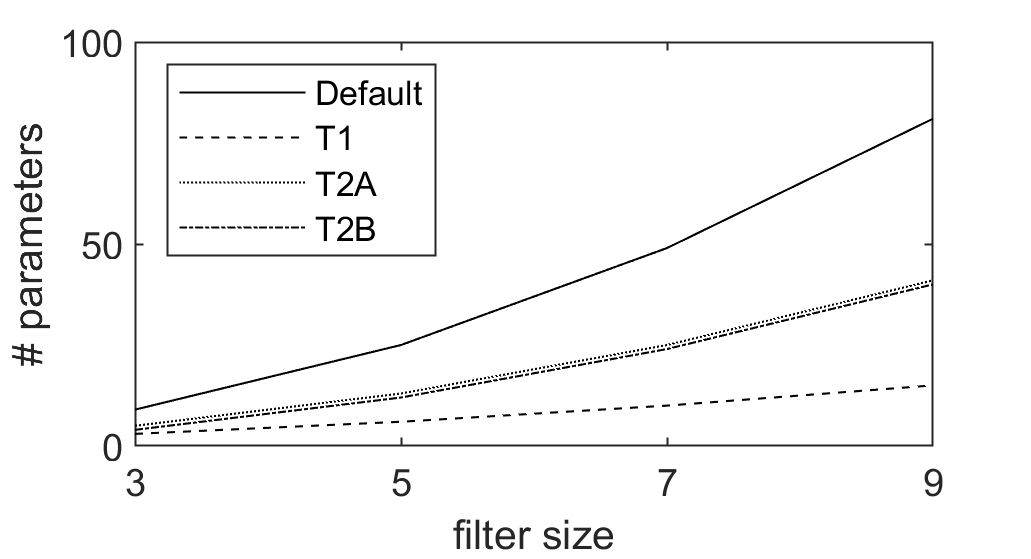}
	\caption{Number of parameters as a function of size the filter size.}
	\label{fig:filtersize}
\end{figure}

The symmetric receptive fields/filters that are introduced in this section corresponds to 2D FIR filters of Type I and III defined in the previous sections. Symmetric Filter Type I (T1) is reminiscent of a Gaussian\textbackslash Laplacian kernel commonly used in image processing. This filter is symmetrical across multiple axes: its center vertically, its center horizontally, and diagonally at each corner. Moreover, filter T1 is capable of teasing out information about point reflection for objects that have central symmetry. T1 can learn Gaussian, Laplacian, Difference of Gaussian types of filters. T2 filters allow to take into account multiple orientations. Symmetric Filter Type 2 (T2) is split into two different filters. We denote by T2A a filter that possesses the property of point reflection. We denote by T2B a filter that possesses the property of anti point reflection, due to the introduction of a negative sign in its second half. The Sobel operator in the $x$ and $y$ dimensions can be learned through T2B only. Filters T1 and T2A have a linear phase due to their symmetry while T2B has a phase onset (antisymmetric coefficients). The number of parameters in T1 and T2 filters compared to default filters are depicted in Fig.~\ref{fig:filtersize}, illustrating the potential gain of free parameters when there exists a large number of filters. In addition to the reduction of the number of parameters to learn, symmetric filters decrease the complexity of the forward operation for estimating the outputs in the CNN: the inputs can be first summed before being multiplied by the weights.

The weights of the filters are defined as follows:
\begin{eqnarray}
T1(x-x_c,y-y_c) & = & T1(x \pm x_c,y \pm y_c) \\
T2A(x-x_c,y-y_c) & = & T2A(x+x_c,y+y_c) \\
T2B(x-x_c,y-y_c) & = & -T2B(x+x_c,y+y_c) \\
T2B(x_c,y_c) & = & 0
\end{eqnarray}
where $(x_c,y_c)$ represents the center of the filter, $1 \leq x,y \leq N_w$.

For the total number of different weights for T1, T2A, and T2B is:
\begin{eqnarray}
N_{T1} & = & (N_{half}+1)(N_{half}+2)/2 \\
N_{T2A} & = & N_{half} \cdot N_w  + N_{half} + 1 \\
N_{T2B} & = & N_{half} \cdot N_w  + N_{half}
\end{eqnarray}
For the forward operation, for each filter, the number of multiplications decreases from $N_w^2$ to $N_{T1}$, $N_{T2A}$ or $N_{T2A}$.

\setlength{\tabcolsep}{4pt}

\begin{figure*}
\centering
\begin{tabular}{@{}c@{}c@{}c@{}c@{}}
$\left[ \begin{tabular}{@{}ccccc@{}}
$w_{1}$ & $w_{2}$ & $w_{3}$ & $w_{4}$ & $w_{5}$ \\
$w_{6}$ & $w_{7}$ & $w_{8}$ & $w_{9}$ & $w_{10}$ \\
$w_{11}$ & $w_{12}$ & $w_{13}$ & $w_{14}$ & $w_{15}$ \\
$w_{16}$ & $w_{17}$ & $w_{18}$ & $w_{19}$ & $w_{20}$ \\
$w_{21}$ & $w_{22}$ & $w_{23}$ & $w_{24}$ & $w_{25}$ \\ 
\end{tabular}  \right]$ &
$\left[ \begin{tabular}{@{}ccccc@{}}
$w_{1}$ & $w_{2}$ & $w_{3}$ & $w_{2}$ & $w_{1}$ \\
$w_{2}$ & $w_{4}$ & $w_{5}$ & $w_{4}$ & $w_{2}$ \\
$w_{3}$ & $w_{5}$ & $w_{6}$ & $w_{5}$ & $w_{3}$ \\
$w_{2}$ & $w_{4}$ & $w_{5}$ & $w_{4}$ & $w_{2}$ \\
$w_{1}$ & $w_{2}$ & $w_{3}$ & $w_{2}$ & $w_{1}$ \\ 
\end{tabular} \right]$ &
$\left[ \begin{tabular}{@{}ccccc@{}}
$w_{1}$ & $w_{2}$ & $w_{3}$ & $w_{4}$ & $w_{5}$ \\
$w_{6}$ & $w_{7}$ & $w_{8}$ & $w_{9}$ & $w_{10}$ \\
$w_{11}$ & $w_{12}$ & $w_{13}$ & $w_{12}$ & $w_{11}$ \\
$w_{10}$ & $w_{9}$ & $w_{8}$ & $w_{7}$ & $w_{6}$ \\
$w_{5}$ & $w_{4}$ & $w_{3}$ & $w_{2}$ & $w_{1}$ \\ 
\end{tabular} \right]$ &
$\left[ \begin{tabular}{@{}ccccc@{}}
$w_{1}$ & $w_{2}$ & $w_{3}$ & $w_{4}$ & $w_{5}$ \\
$w_{6}$ & $w_{7}$ & $w_{8}$ & $w_{9}$ & $w_{10}$ \\
$w_{11}$ & $w_{12}$ & 0 & $-w_{12}$ & $-w_{11}$ \\
$-w_{10}$ & $-w_{9}$ & $-w_{8}$ & $-w_{7}$ & $-w_{6}$ \\
$-w_{5}$ & $-w_{4}$ & $-w_{3}$ & $-w_{2}$ & $-w_{1}$ \\  
\end{tabular} \right]$ \\
R (default) & T1 & T2A & T2B
\end{tabular}
\caption{The four types of filters: R, T1, T2A, and T2B for a filter of size $5 \times 5$.}
\label{fig:filters}
\end{figure*}

\setlength{\tabcolsep}{6pt}

All filters are initialized by randomly sampling a normal distribution with zero mean and standard deviation inversely proportional to the square root of the fan in to a neuron at some given layer~\cite{LeCun2012}. Filter T1 only requires 6 weights to be generated, in the case of a $ 5 \times 5 $ filter, which then get copied to their appropriate positions. T2A and T2B require 13 and 12 weights, respectively. Without the symmetrical constraint, there are 25 free parameters to tune throughout the training process in the case of a single $5 \times 5$ filter.
The weights for the default condition (R), T1, T2A, and T2B are presented in Fig.~\ref{fig:filters}. We denote by $w_{i,j}$ the value of the weight at the position $(i,j)$ in the filter, and $w_{k}$, the $k^{th}$ different weight in the filter. For instance, for T1 we have $w_{1}=w_{1,1}=w_{N_w,1}=w_{1,N_w}=w_{N_w,N_w}$.

In order to insure that the filters retain symmetry throughout the training process, it is necessary to modify a portion of the backpropagation algorithm. The same way that weight sharing is achieved through averaging the errors on all the connections that share the same weight, it is necessary to share the weight between the elements of the filter that have the same weight.
Within the weight update procedure, after the gradients for the receptive fields have been computed, they are not directly added to the current weights. Instead the gradients are passed off to the specific weight update procedure for the filter being used within a given layer. 

The update operation was experimented to determine what would yield the best results. Initially, an averaging operation was executed by summing the gradients for each weight with its symmetric counterparts. However, this was determined to decrease the gradient too much. Instead, the sum of the gradients was considered and implemented. To give a much more general description of the update procedures, let $i$ and $j$ index the rows and columns of a 2D filter, let $\Delta \mathbf{W}$ be a matrix of gradients for a 2D filter. For a filter of size $N_w \times N_w$ where $N_w$ is odd, the updates proceed as follows.
\begin{eqnarray}
\Delta \mathbf{W} & = &  
\begin{bmatrix}
\delta_{1,1} &  \cdots & \delta_{1,N_b} \\
\vdots       &  \vdots & \vdots \\
\delta_{N_b,1} & \cdots & \delta_{1,N_b}  \\
\end{bmatrix}
\label{eq:grad-mat}
\end{eqnarray}

For T1, we define the distance of an element $(i,j)$ from the central element $(N_{half}+1,N_{half}+1)$ of the filter as
\begin{align}
	d(i,j) = \sqrt{ \left( i - \left( N_{half} + 1 \right) \right)^{2} +  \left( j - \left( N_{half} + 1 \right) \right)^{2} }.
\end{align}

Let $S_k$ be the set of all elements defined by their coordinate $(i,j)$ in the filter that are at the same distance away from the center element.
\begin{eqnarray}
\forall ((i_1,j_1),(i_2,j_2)) \in S_k^2 & \rightarrow & d(i_1,j_1)=d(i_2,j_2) \\
|S_k| & = & N_{T1} \nonumber
\end{eqnarray}
with the weight $w_k$ associated to each set $S_k$, $1 \leq k \leq N_{T1}$.  

For the propagation step, the output of the convolution is defined by:
\begin{eqnarray}
y_{T1}(i,j)  & = & \sum\limits_{k=1}^{N_{T1}} \left( w_{k} \cdot \sum_{(i_1,j_1) \in S_{k}} x(i_1,j_1) \right) \\
y_{T2A}(i,j) & = &\sum\limits_{(i_1,j_1) \in S_{T2A}}  w_{i_1,j_1} \cdot \nonumber \\
             &   &  (x(i+i_1,j+j_1)+x(i+i_2,j+j_2)) \nonumber \\
y_{T2B}(i,j) & = & \sum\limits_{(i_1,j_1) \in S_{T2B}} w_{i_1,j_1} \cdot \nonumber \\
             &   &   (x(i+i_1,j+j_1)-x(i+i_2,j+j_2)) \nonumber             
\end{eqnarray}
where $i3=N-i1+1$ and $j3==N-j1+1$ for $y_{T2A}$ and $y_{T2B}$. $S_{T2A}$ and $S_{T2B}$ represent the set of coordinates containing different weights for T2A and T2B, respectively. All the expressions are equivalent to the original convolution operation but the number of multiplications is reduced due to the shared weights within the filter.

\subsubsection{T1 Generalized Update Procedure}

The gradient for the weight $b_k$, $1 \leq k \leq N_{T1}$, within the filter is computed as follows:
\begin{eqnarray}
\delta_{\mathbf{T1}[k]} & = & \sum_{(i,j) \in S_k}{\Delta \mathbf{W}[i,j]}.
\end{eqnarray}

\subsubsection{T2A Generalized Update Procedure}

For T2A the gradient update for the weight $b_{i,j}$ within the filter is computed as follows:
\begin{eqnarray}
	\delta_{\mathbf{T2A}[i,j]} & = & \Delta \mathbf{W}[i,j] + \\
	                           &   & \Delta \mathbf{W}[N_w - (i-1), N_w - (j-1)]. \nonumber
\end{eqnarray}

\subsubsection{T2B Generalized Update Procedure}

For T2B the gradient update for a weight at location $(i,j)$ within the filter as long as  $(i,j) < (N_{half} + 1, N_{half} + 1 )$ (positive weights in in T2B) is computed as follows:
\begin{eqnarray}
\delta_{\mathbf{T2B}[i,j]} & =  & \Delta \mathbf{W}[i,j] \\
\delta_{\mathbf{T2B}[ N_w - (i-1), N_w - (j-1)]} & = & - \mathbf{T2B}[i,j] \nonumber
\end{eqnarray}

The positive weights in filter $\mathbf{T2B}$ are simply updated with their appropriate gradients in $\mathbf{W}$. Then these new positive weights are negated and copied over into the negative half of $\mathbf{T2B}$.

\subsubsection{Center Element Update Procedure}

For all filters, except T2B, the center element is updated as follows:
\begin{align}
	\delta_{\mathbf{T_{any}}\left[ N_{half} + 1 , N_{half} + 1 \right]} = \Delta \mathbf{W} \left[N_{half} + 1 , N_{half} + 1 \right]
\end{align}
For T2B, the center element is never updated, it is initialized and remains zero throughout training.

\section{Database Descriptions}
\label{sec:database-descriptions}

We consider four datasets corresponding to handwritten numerals of different scripts and in which it is possible to consider the same CNN architecture, so it is possible to focus on the differences related to the type of filters. These databases were chosen as they have different numbers of training examples. 

The MNIST database contains digits of the Latin script, it is a benchmark in supervised classifiers~\cite{lecun1998,simard2003}. It includes a training and test database of 60000 and 10000 images, respectively. For the comparison of techniques, two characteristics are typically precised if they are used or not: the addition of distorted images in the database, and the type of normalization of the images. The error rate reaches quasi human performance level of 0.23\% with a combination of 35 convolutional neural networks, where each network requires almost 14h to be trained~\cite{ciresan2012}. In addition, we consider three datasets of handwritten Indian numerals: Bangla~\cite{chaud2006,vajda2009}, Devanagari, and Oriya. These databases were created at the Indian Statistical Institute, Kolkata, India~\cite{chaud1998,pal2004,bha2005}. The second database corresponds to Devanagari digits, which is part of the Brahmic family of scripts of India, Nepal, Tibet, and South-East Asia~\cite{sethi1977}. The Oriya script is one of the many descendants of the Brahmi script of ancient India. In~\cite{bhowmick2006}, Bhowmik et al. obtain an accuracy of 90.50\% by using Hidden Markov Models.

The images were preprocessed the same way as in the original images in the MNIST database. Because some databases have noisy images and/or images in color, images were first binarized with the Otsu method at their original size~\cite{otsu1979}, then they were size normalized to fit in a 20 $\times$ 20 pixel box while preserving their aspect ratio. The resulting images contain 8 bit gray levels due to the bicubic interpolation for resizing the images. Finally, all the images were centered in a 28 $\times$ 28 pixel box field by computing the center of mass of the pixels, and translating the gravity center of the image to the center of the 28 $\times$ 28 field. Finally, an additional 1 pixel border is added to the top and left side of every image to change the size to $29 \times 29$ to fit into the CNN architecture. The total number of images and the number of images per class for each dataset is presented in Table~\ref{table:database}. Images in all datasets from the training files were split, with 90\% in a training set and rest in a validation set. The images were z-score normalized, i.e., by removing the mean and divided by the standard deviation across all examples in the training dataset.

\begin{table}
\caption{Data distribution.}
\label{table:database}
\centering
\begin{tabular}{@{}c@{}|cccc}
Database    & MNIST & Bangla & Devanagari & Oriya \\ \hline
& \multicolumn{4}{c}{\textbf{Training}} \\
\# samples & 60000 & 19392 & 18783 & 4970 \\
\# per class & $6000\pm339$ & $1939\pm9$ & $1878\pm15$ & $497\pm3$ \\
& \multicolumn{4}{c}{\textbf{Test}}  \\
\# samples & 10000 & 4000 & 3763 & 1000 \\
\# per class & $1000\pm62$ & 400  & $376\pm3$ & 100 \\
\end{tabular}
\end{table}

\begin{figure}[ht]
\centering
\begin{tabular}{cc}
\subfigure[Western Arabic (Latin script)]{\includegraphics[width=0.45\columnwidth]{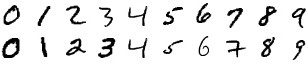}\label{fig:mnist}} &
\subfigure[Bangla]{\includegraphics[width=0.45\columnwidth]{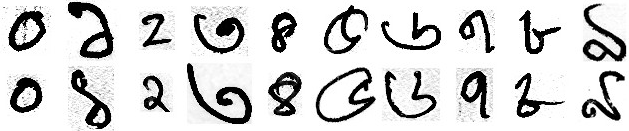}\label{fig:bangla}} \\ 
\subfigure[Devanagari]{\includegraphics[width=0.45\columnwidth]{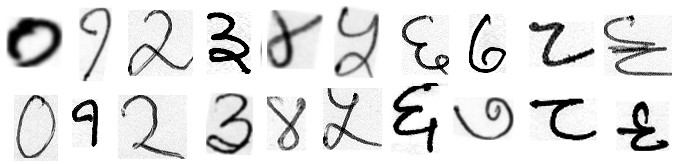}\label{fig:devnagari}} & 
\subfigure[Oriya]{\includegraphics[width=0.45\columnwidth]{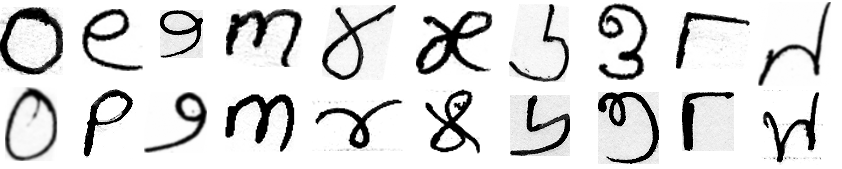}\label{fig:oriya}}
\end{tabular}
\caption{Representative handwritten digits for the different databases (from zero to nine).}
\label{fig:digits}
\end{figure}

\begin{table*}[th!]
	\centering
	\begin{tabular}{|l c|c|c|c|c|c|c|c|c|}
		\hline
		\multicolumn{2}{|c|}{Network} & \multicolumn{2}{|c|}{MNIST} & \multicolumn{2}{|c|}{Bangla} &  \multicolumn{2}{|c|}{Devanagari} & \multicolumn{2}{|c|}{Oriya}\\
		\hline
		& & Acc & CE & Acc & CE & Acc & CE & Acc & CE \\ \hline
		L-R-R            & &$98.23\pm 0.09$  &  $0.064$& $96.64\pm 0.15$ &  $0.125$& $97.00 \pm 0.10$ & $0.122$ & $95.52\pm0.26$ & $0.168$\\ 
		L-T1-T1          & &$97.21\pm 0.16$  &  $0.105$& $94.33\pm 0.24$ &  $0.209$& $95.61 \pm 0.23$ & $0.173$ & $94.34\pm0.36$ & $0.216$\\
		L-T2A-T2A        & &$98.00\pm 0.11$  &  $0.070$& $95.97\pm 0.15$ &  $0.143$& $96.79 \pm 0.13$* & $0.125$ & $94.72\pm0.16$ & $0.193$\\ 
		L-T2B-T2B        & &$97.89\pm 0.18$  &  $0.078$& $96.21\pm 0.12$ &  $0.142$& $96.67 \pm 0.14$ & $0.127$ & $95.46\pm0.23$\textbf{*} & $0.168$\\ 
		L-T1-R           & &$97.93\pm 0.10$  &  $0.078$& $95.84\pm 0.22$ &  $0.155$& $96.38 \pm 0.13$ & $0.143$ & $95.26\pm0.24$\textbf{*} & $0.168$\\ 
		L-T2A-R          & &$98.19\pm 0.05$\textbf{*} &  $0.065$& $96.28\pm 0.11$ &  $0.140$& $96.98 \pm 0.18$\textbf{*} & $0.120$ & $95.50\pm0.06$\textbf{*} & $0.170$\\ 
		L-T2B-R          & &$98.09\pm 0.10$\textbf{*} &  $0.069$& $96.26\pm 0.27$\textbf{*} &  $0.137$& $96.74 \pm 0.07$ & $0.120$ & $95.42\pm0.34$\textbf{*} & $0.171$\\ 
		\hline
		F-R-R        		& &$93.78\pm5.25$  & $0.201$& $89.67\pm 4.01 $ & $0.334$& $88.31\pm9.65$  & $0.478$ & $88.76\pm2.36$ & $0.453$\\ 
		F-T1-T1      		& &$96.35\pm0.27$  & $0.121$& $71.29\pm 24.63$ & $0.865$& $67.81\pm31.31$ & $0.900$ & $90.50\pm4.65$ & $0.326$\\ 
		F-T2A-T2A    		& &$84.39\pm20.55$ & $0.459$& $62.44\pm 22.73$ & $1.134$& $77.41\pm15.35$ & $0.837$ & $83.16\pm9.04$ & $0.719$\\ 
		F-T2B-T2B    		& &$91.65\pm3.81$  & $0.291$& $85.24\pm  3.51$ & $0.497$& $87.99\pm4.03$  & $0.417$ & $87.88\pm2.48$ & $0.425$\\ 
		F-T1-R       		& &$86.46\pm19.06$ & $0.398$& $88.85\pm  5.26$ & $0.367$& $81.63\pm18.71$ & $0.567$ & $91.36\pm3.21$ & $0.328$\\ 
		F-T2A-R      		& &$95.10\pm 2.10$ & $0.166$& $76.41\pm 17.26$ & $0.795$& $80.47\pm7.02$  & $0.794$ & $90.86\pm1.57$ & $0.340$\\ 
		F-T2B-R      		& &$91.83\pm 3.93$ & $0.306$& $88.80\pm  3.22$ & $0.371$& $87.19\pm4.29$  & $0.477$ & $90.82\pm0.78$ & $0.300$\\ 
		\hline
	\end{tabular}
	\caption{Performance for all datasets. Test set accuracy and cross entropy, averaged across five different runs, for each of the filter combinations for learned (`L') and fixed (`F') networks. The $*$ represents results with no significant difference with \textsl{L-R-R}.}
	\label{table:DatasetResults}
\end{table*}

\section{Architecture}
\label{sec:architecture}

The CNN architecture chosen for testing consisted of 4 layers, where the first two layers are convolutions and the last two layers are fully connected. The architecture is based on a state of the art architecture that does not require a large number of layers~\cite{simard2003}. This architecture was chosen as it is possible to decompose the network into two clear stages: feature extraction through 2 convolutional layers~\cite{jarrett2009}, and classification with 1 fully connected hidden layer. Fig.~\ref{fig:testarch} depicts the network along with several other parameters that were kept constant, such as the activations used and layer sizes.

\begin{center}
	\centering
	\includegraphics[width=5cm]{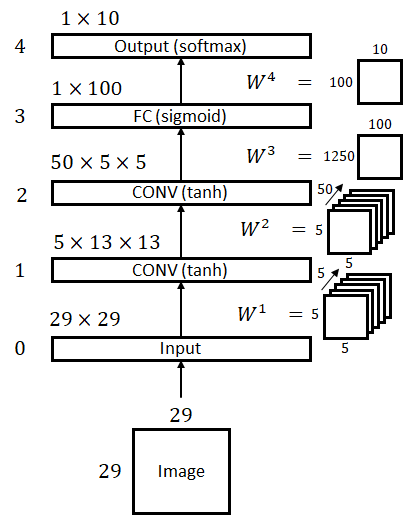}
	\captionof{figure}{Architecture of the CNN.}
	\label{fig:testarch}
\end{center}

Additionally, the network learning rate was set to 0.001 with no optimizers in use and no scheduled learning rate decreases. However, the network does scale the learning rate for each layer depending on the fan-in for a particular neuron of that layer~\cite{LeCun2012}. The weights are randomly sampled from a Gaussian distribution with mean $\mu = 0$ and standard deviation $\sigma = \frac{1}{\sqrt{m}}$, where $m$ is the fan in for a given layer~\cite{LeCun2012}. The loss function used was cross entropy (CE).	

For performance evaluation, we compare 14 conditions decomposed into two main types: `L' (learned) and `F' (fixed). In the learned conditions, all the convolutional kernels are learned while in the fixed condition, we consider fixed kernels that are initialized randomly, following the principle of the random projections used in the extreme learning machine (ELM) paradigm~\cite{huang2006,huang2015}. For each type, we estimate the performance of different types of convolutional layers (R for the default case, T1, T2A, and T2B) in the two first convolutional layers. For instance the condition \textsl{L-T1-R} corresponds to learned convolutional layers, with T1 for the first convolutional layer and R for the second convolutional layer. For assessing the differences across conditions, we consider a Wilcoxon rank sum test. If there is a failure to reject the null hypothesis at the 5\% significance level, we consider that two conditions are equivalent. In particular, we focus the comparisons on \textsl{L-R-R} versus the other `L' conditions.

\section{Results}
\label{sec:results}

\begin{figure*}[th!]
\centering	
\begin{tabular}{@{}c@{}c@{}c@{}c@{}}	
\includegraphics[width=4.4cm]{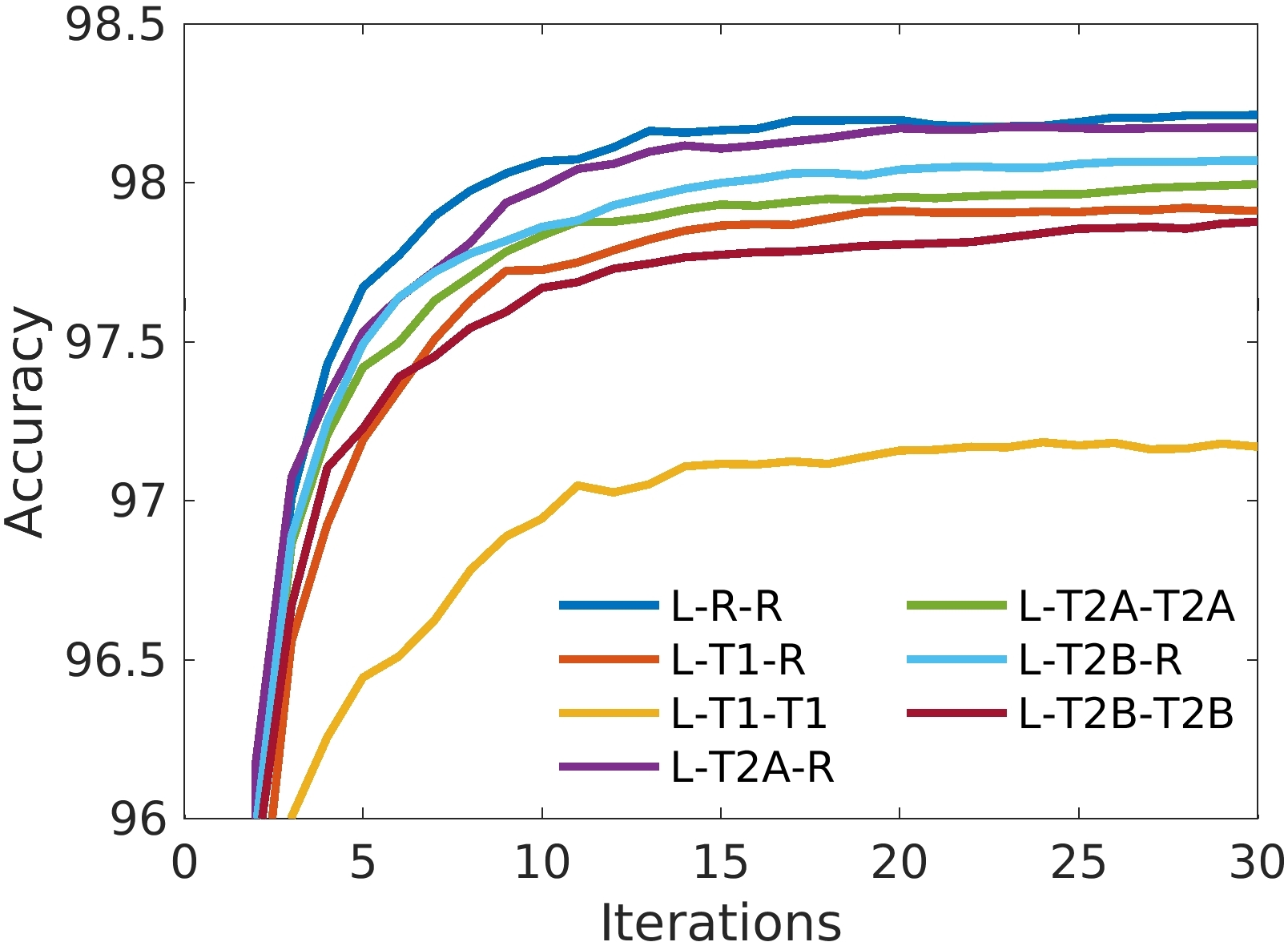} &
\includegraphics[width=4.4cm]{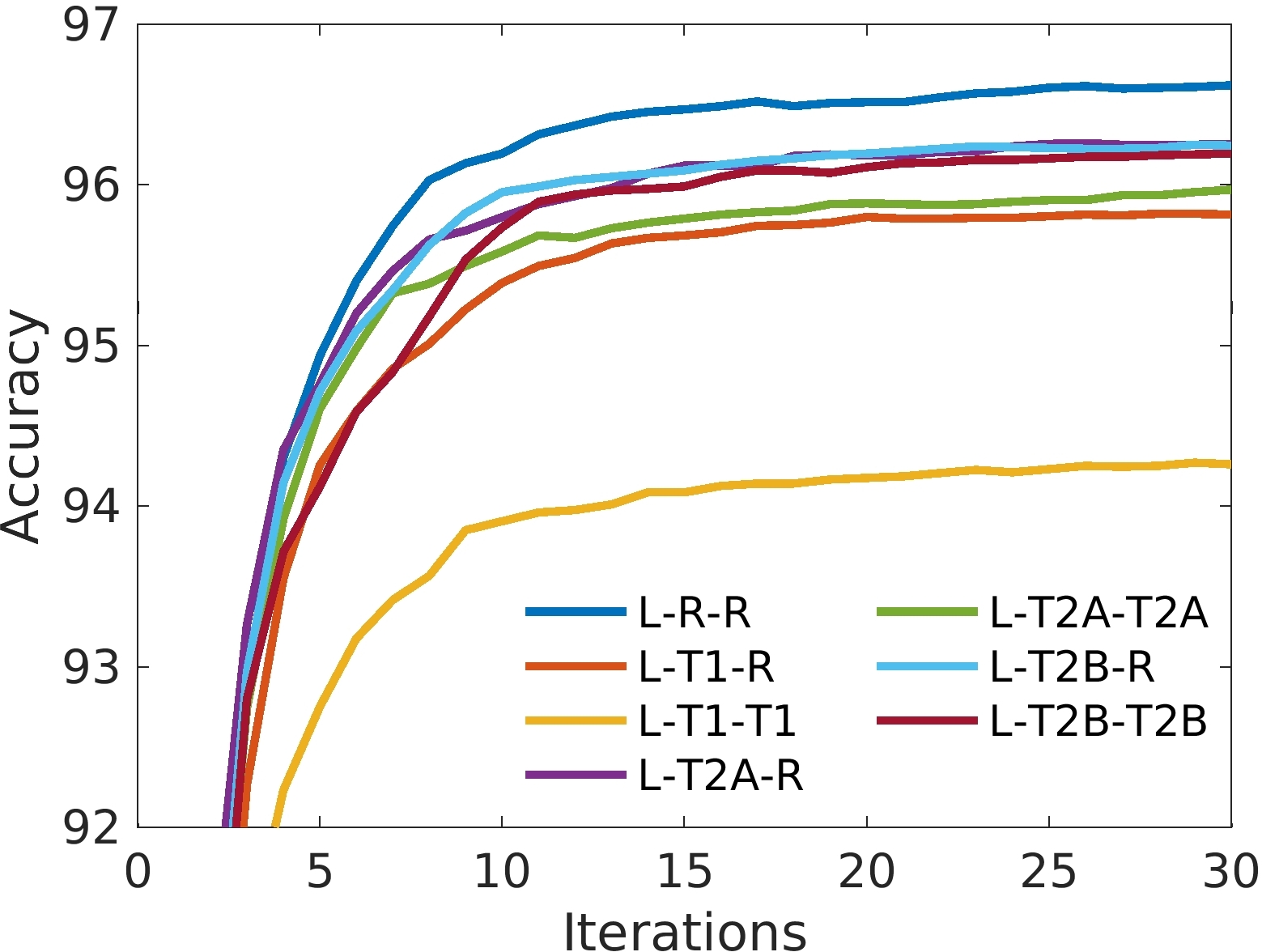} &
\includegraphics[width=4.4cm]{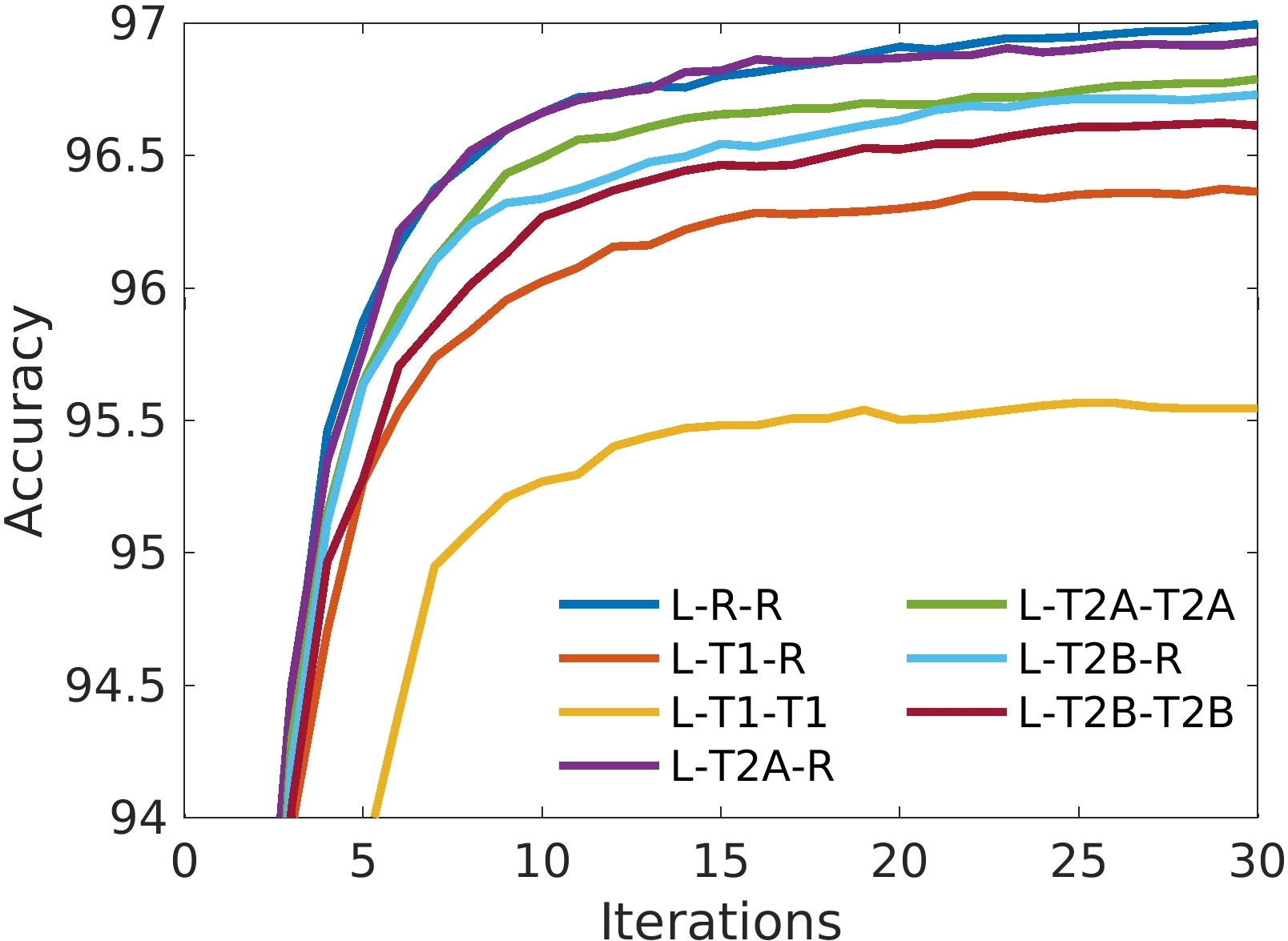} &
\includegraphics[width=4.4cm]{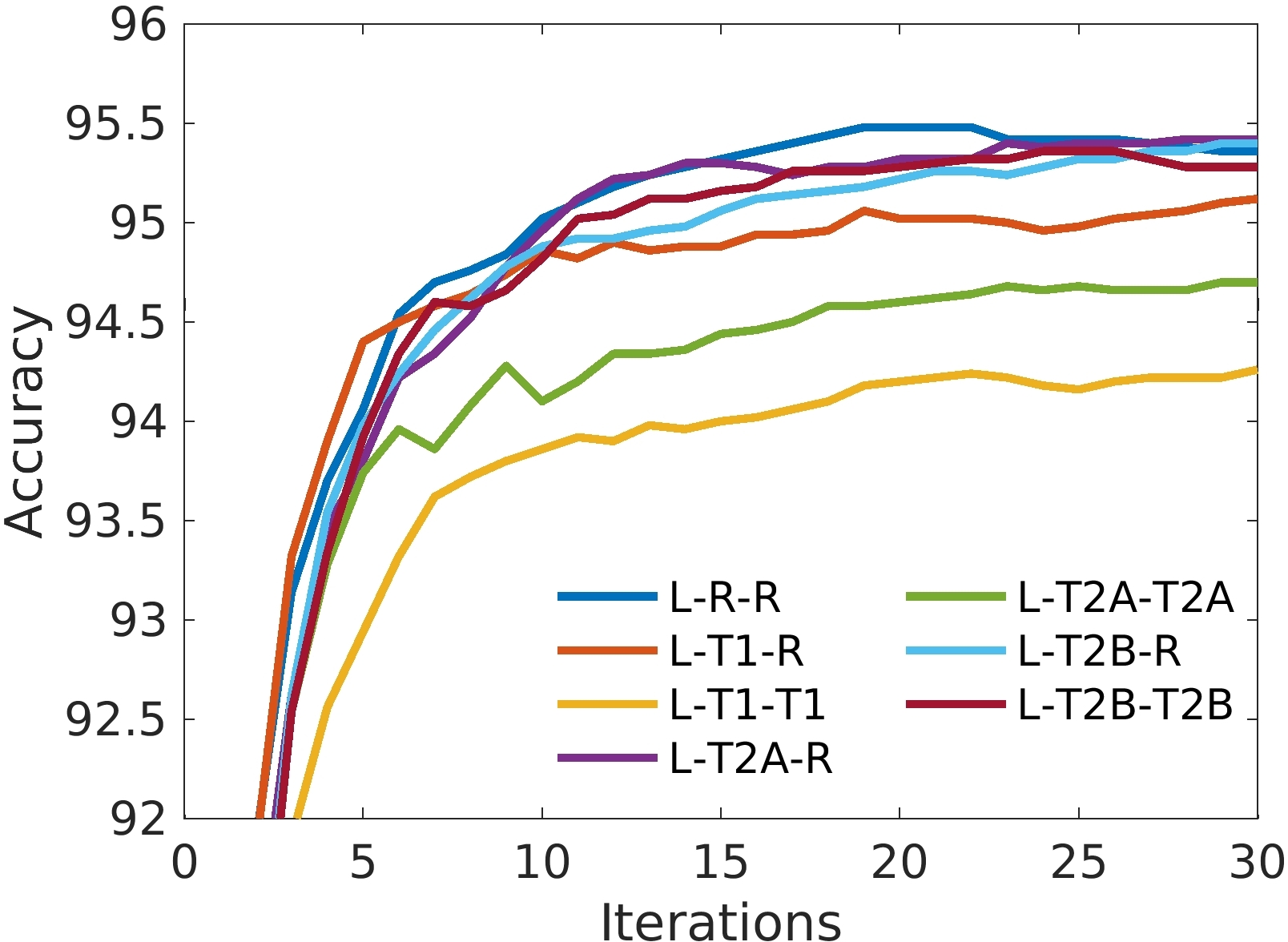} \\
\includegraphics[width=4.4cm]{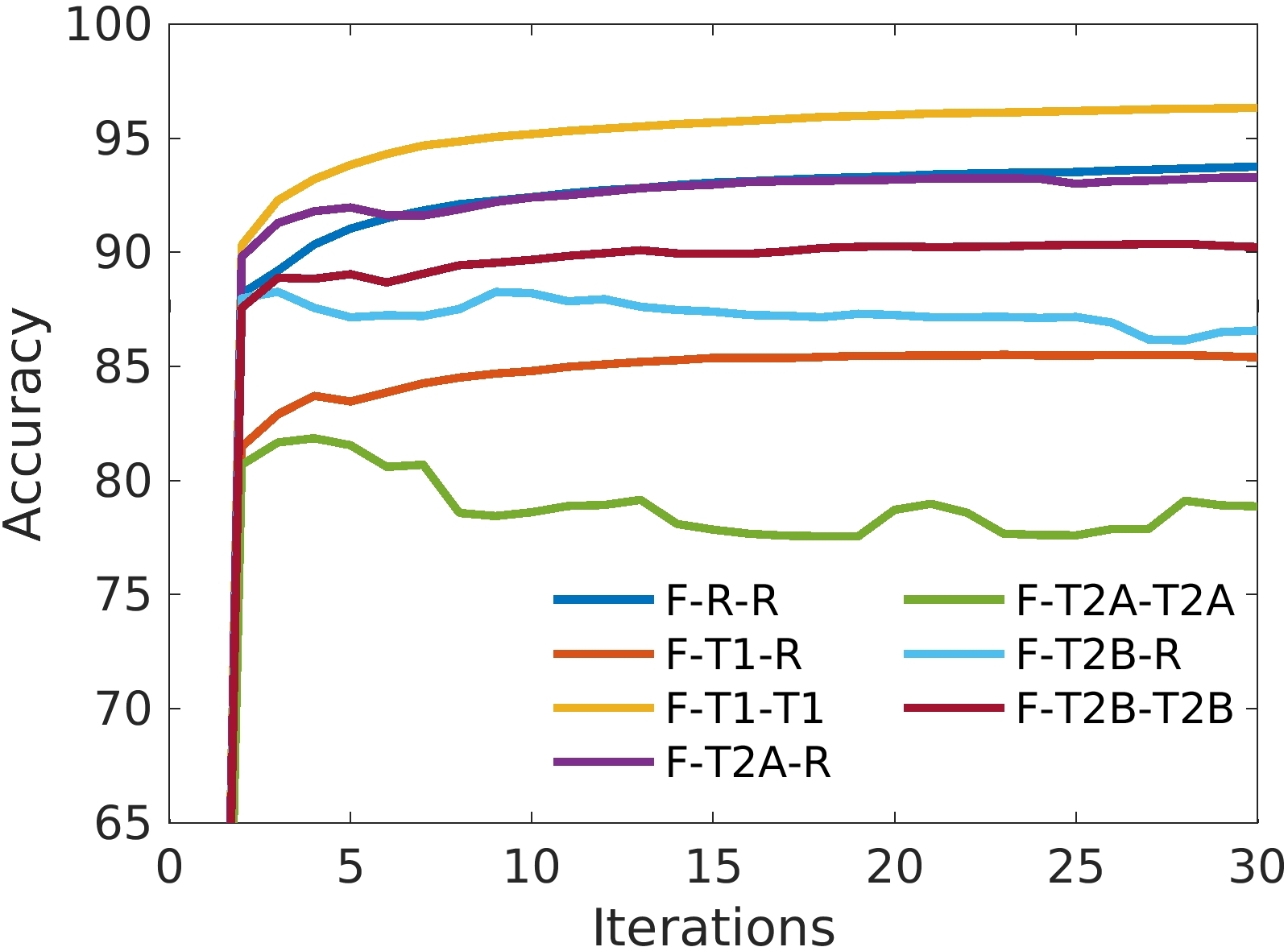} &
\includegraphics[width=4.4cm]{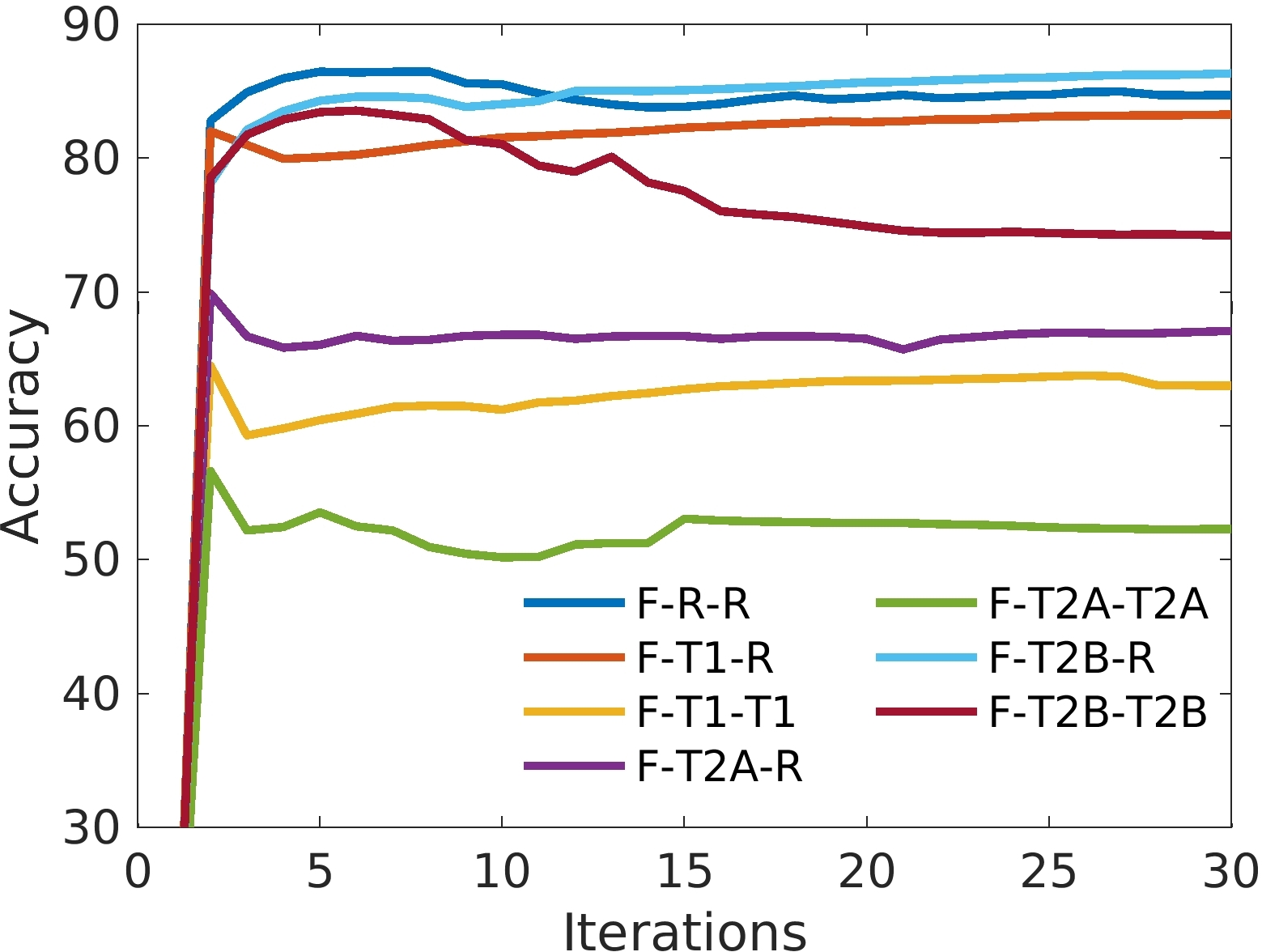} &
\includegraphics[width=4.4cm]{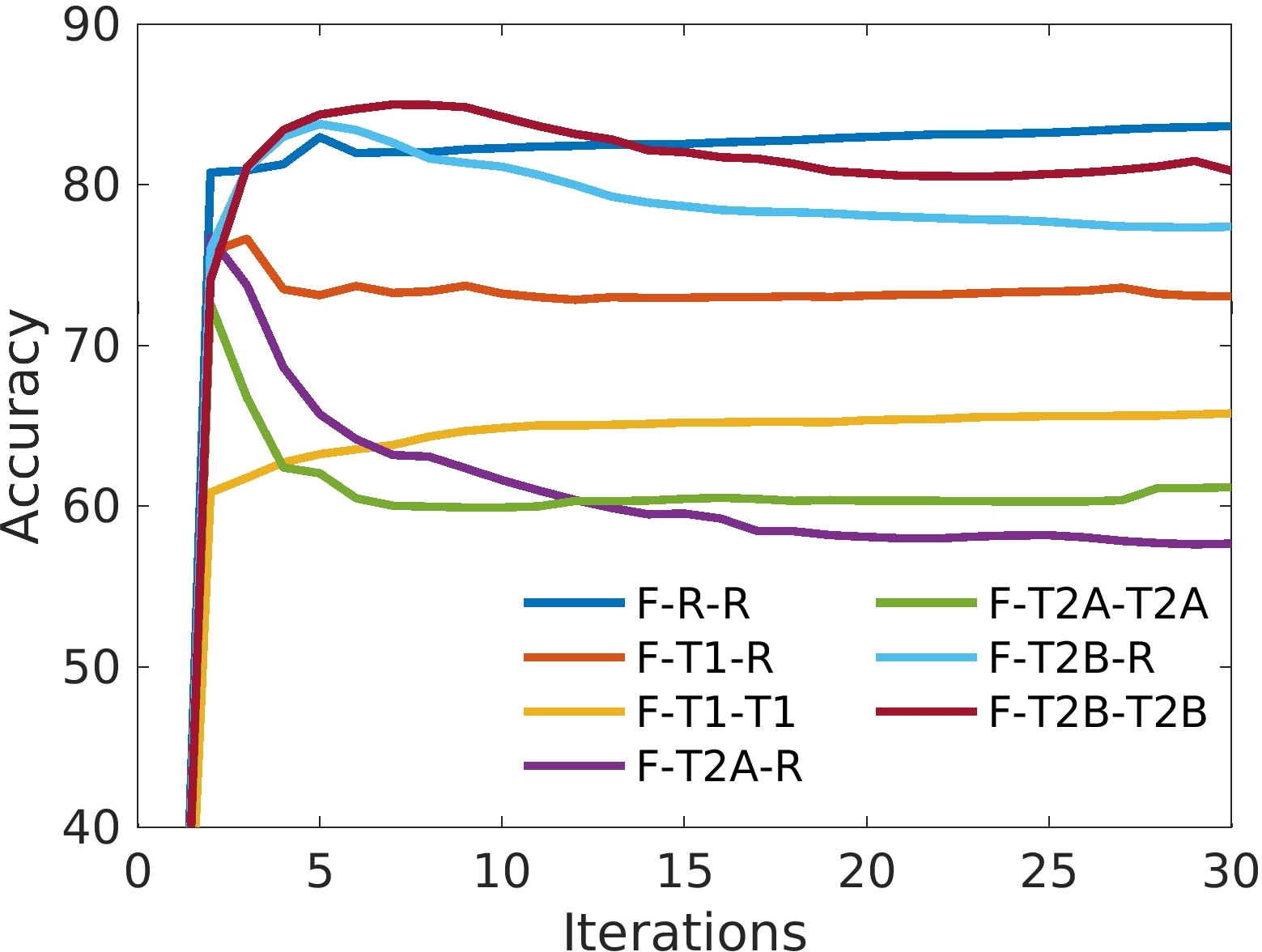} &
\includegraphics[width=4.4cm]{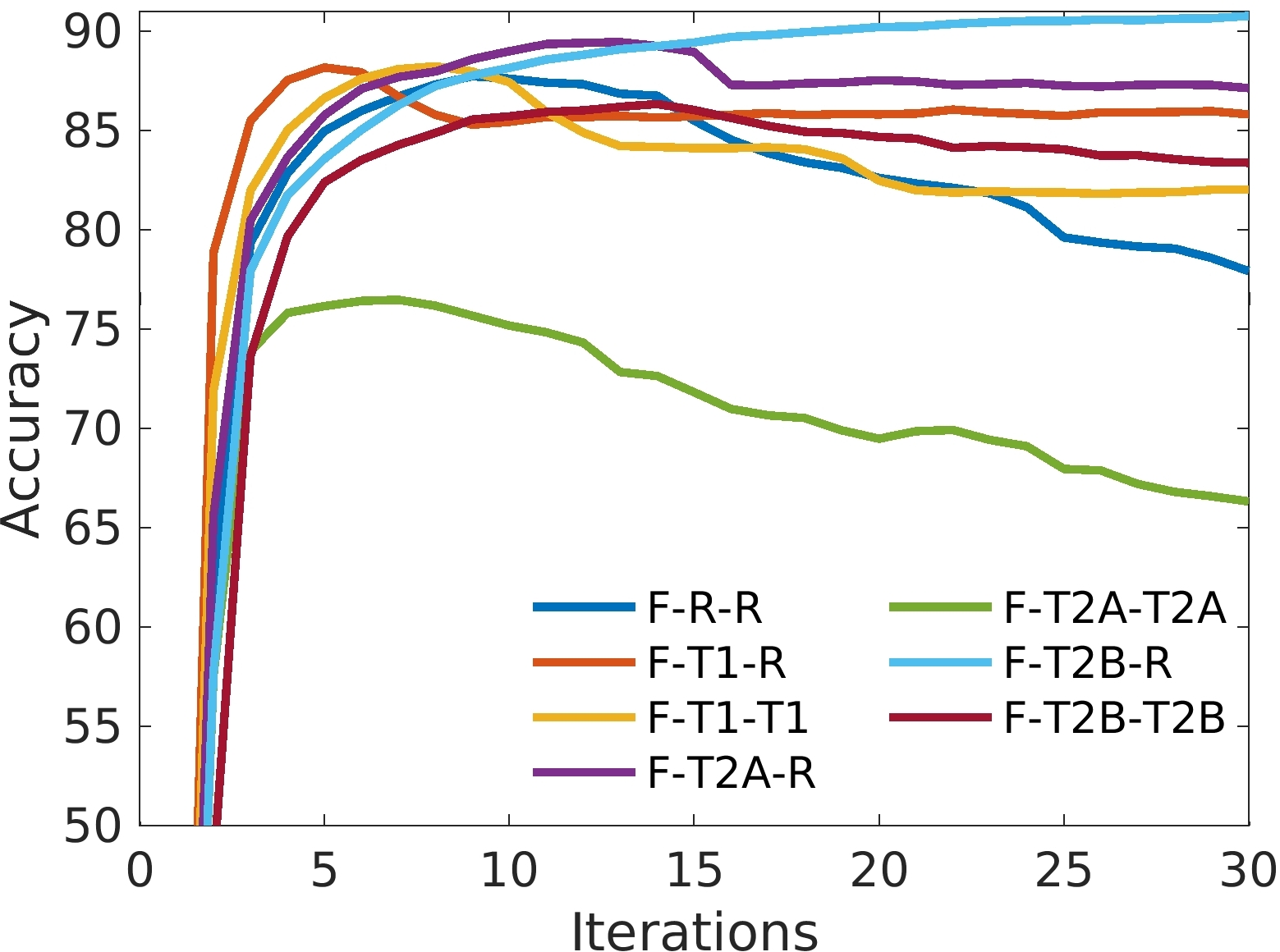} \\
MNIST & Bangla & Devanagari & Oriya \\
\end{tabular}
\caption{Test set accuracy, averaged across five runs for each filter combination. \textbf{Top} row: learned (`L') conditions, \textbf{bottom} row: fixed (`F').}
\label{fig:DatasetAccPlots}
\end{figure*}

This section summarizes the results of running tests with several different combinations of the various filter types (i.e R, T1, T2A, and T2B) across the convolutional layers of the network described in Section~\ref{sec:architecture}. Each network with a unique filter combination was trained five times for each dataset listed in Section~\ref{sec:database-descriptions}. All data presented in this section was produced by averaging the test dataset results across the five separately trained networks for a given dataset and filter combination. The baseline, labeled \textsl{R-R} in Table~\ref{table:DatasetResults}, had all filters in both of its convolutional layers set randomly and updated every parameter in each filter. 
The mean and standard deviation of the accuracy (Acc), in \%, and the cross-entropy loss value are presented in Table~\ref{table:DatasetResults} for all the different conditions and datatsets. Graphs of the test performance for each network on every iteration are shown in Fig.~\ref{fig:DatasetAccPlots}.

For MNIST, \textsl{L-R-R} slightly edges out as the best performing, this network achieved a mean accuracy of $98.23 \pm 0.09\%$, followed by the networks \textsl{L-T2A-T2A}, \textsl{L-T2A-R} and \textsl{L-T2B-R} with a performance of $98.00\pm0.11$, $98.19\pm0.05$, and $98.09\pm0.10$. It is worth mentioning that it is better to keep the second convolutional layer to the default condition as we observe a drop of performance from \textsl{L-T1-R} to \textsl{L-T1-T1}, from \textsl{L-T2A-R} to \textsl{L-T2A-T2A}, and \textsl{L-T2B-T2B}. Without learning the convolutional kernels, the best performance is obtained with \textsl{L-T1-T1} with an accuracy of $96.35\pm0.27$. This network contains the less number of free parameters. The statistical test reveals no difference between \textsl{L-R-R} and \textsl{L-T2A-R}, \textsl{L-R-R} and \textsl{L-T2B-R}.
For Bangla, \textsl{L-R-R} preforms the best with $96.64\pm0.15$\% and networks \textsl{L-T2A-R}, \textsl{L-T2B-R}, and {L-T2B-T2B} achieve accuracies fairly close to it, i.e. above 96\%. Network \textsl{L-T1-T1} does the worst, with over a $2\%$ difference from the top performing network, whereas all other have less than a $1\%$ difference. The statistical analysis indicates no difference between \textsl{L-R-R} vs \textsl{L-T2B-R}. 
The best accuracy for the Devanagari dataset was attained by \textsl{L-R-R} with $97.00\pm0.10$, but \textsl{L-T2A-R} performed with a relatively similar performance with $96.98\pm0.18$. Nearly all networks for this dataset come in with under a $0.5\%$ difference in accuracy, except for \textsl{L-T1-T1} and \textsl{L-T1-R}. Interestingly, the condition \textsl{F-R-R} provides the best accuracy for the fixed conditions. The statistical tests show no difference between \textsl{L-R-R} and {L-T2A-T2A}, \textsl{L-R-R} and {L-R-T2A}.  
The maximum accuracy on the Oriya dataset was obtained by network \textsl{L-R-R} with $95.52\pm0.26$ followed by \textsl{L-T2A-R} with $95.50\pm0.06$, which has a lower standard deviation.\textsl{L-T2B-T2B} performed well in this case. Note that all combinations achieve greater than $95\%$ except for \textsl{L-T1-T1} and \textsl{L-T2A-T2A}. Interestingly, the fixed weight networks seem to do relatively well when paired with a symmetric filter in the first layer and a random filter for the second layer on this dataset as the best performance is achieved with \textsl{F-T1-R}. It is highly likely that this is due to the fact that the first layer is actually able to extract some important features that helps to better distinguish between images of varying class. The statistical tests indicate no difference between \textsl{L-R-R} and \textsl{L-T2B-T2B}, \textsl{L-T1-R}, \textsl{L-T2A-R}, \textsl{L-T2B-R}, showing key evidence about the advantages of the proposed filters as the number of parameters for feature extraction is reduced to 660 from 1375 in the default case. 

\begin{figure*}[th!]
\centering
\begin{tabular}{@{}c@{}c@{}c@{}c@{}}	
\subfigure[MNIST]{\includegraphics[trim={0 0 1cm 0},clip,width=0.5\figwidth]{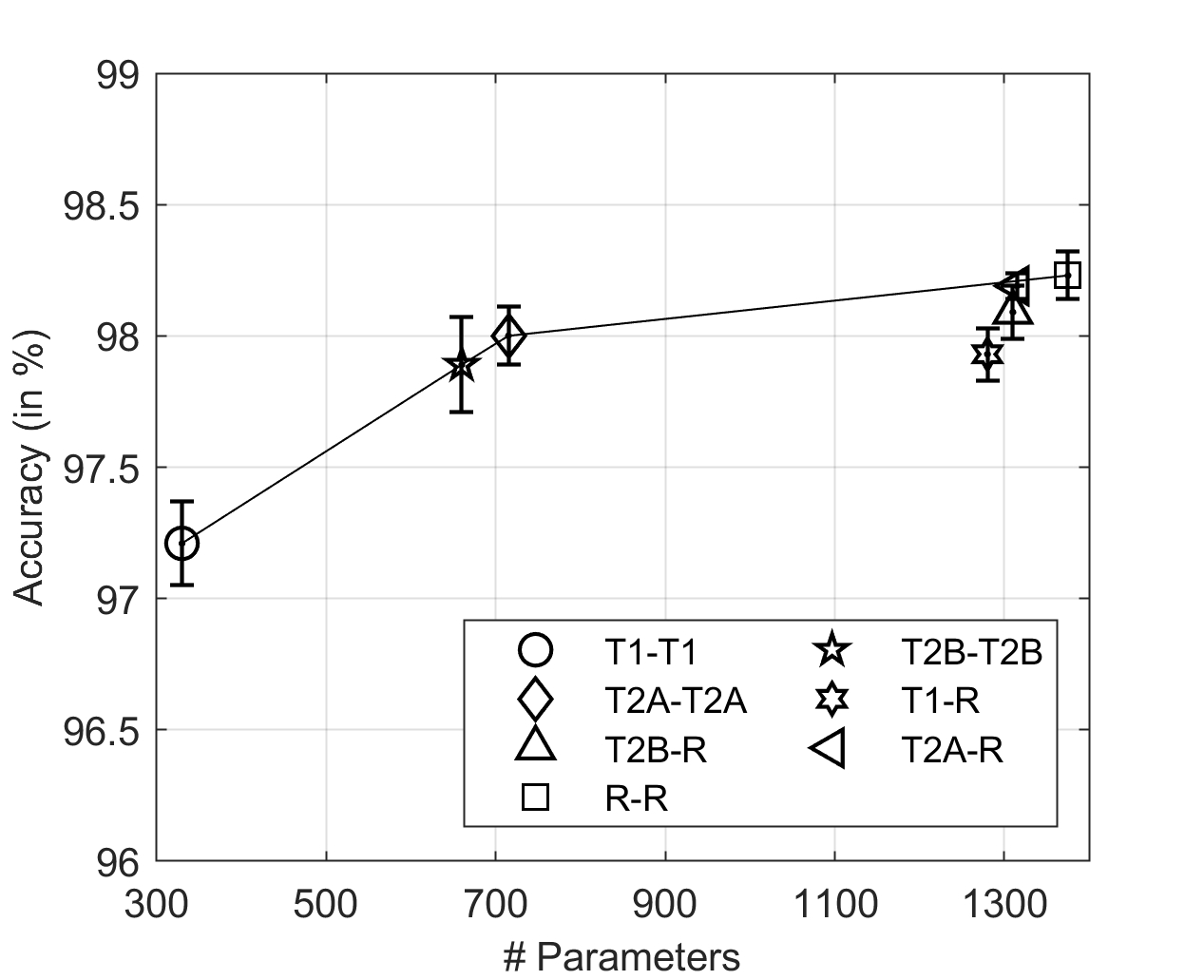}} &
\subfigure[Bangla]{\includegraphics[trim={0 0 1cm 0},clip,width=0.5\figwidth]{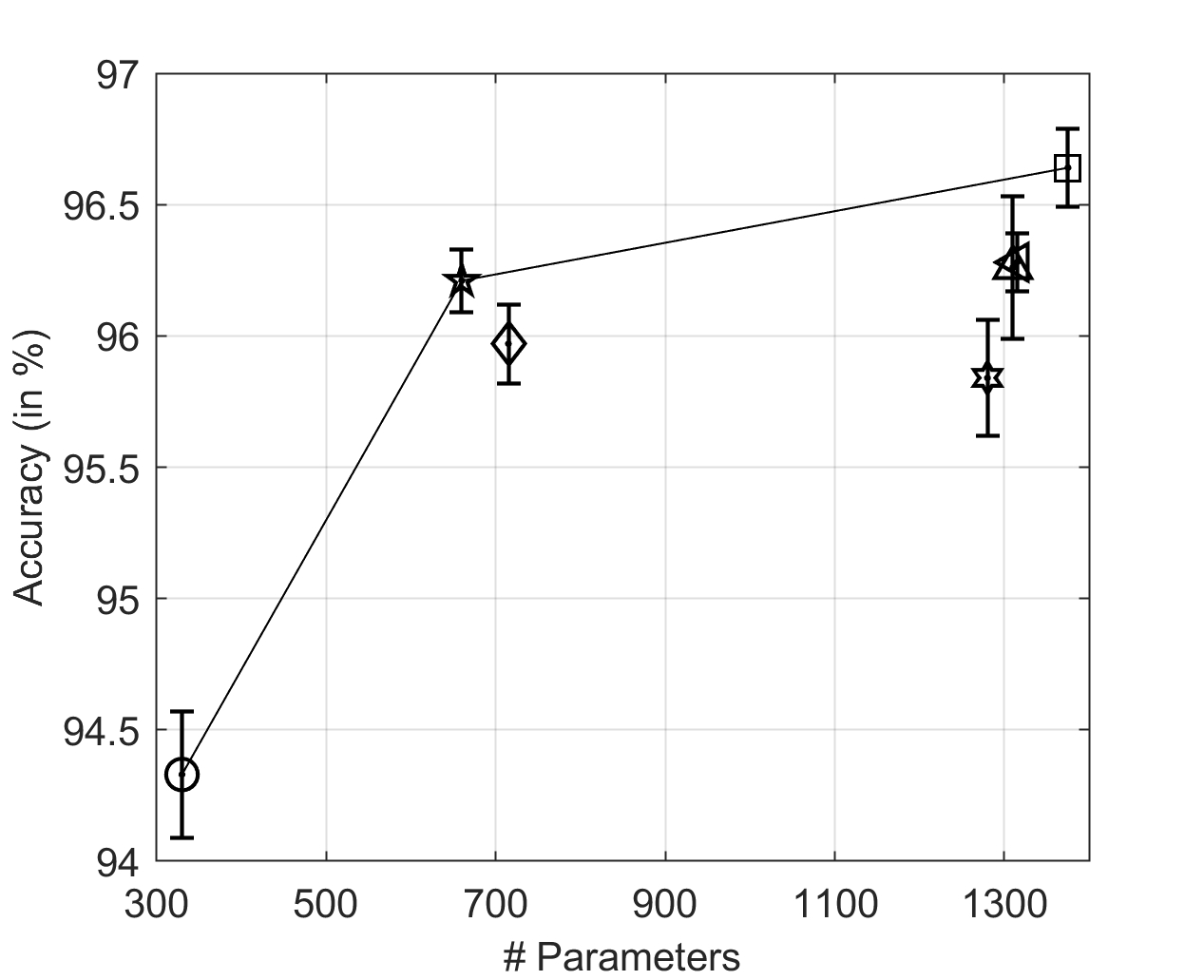}} &
\subfigure[Devanagari]{\includegraphics[trim={0 0 1cm 0},clip,width=0.5\figwidth]{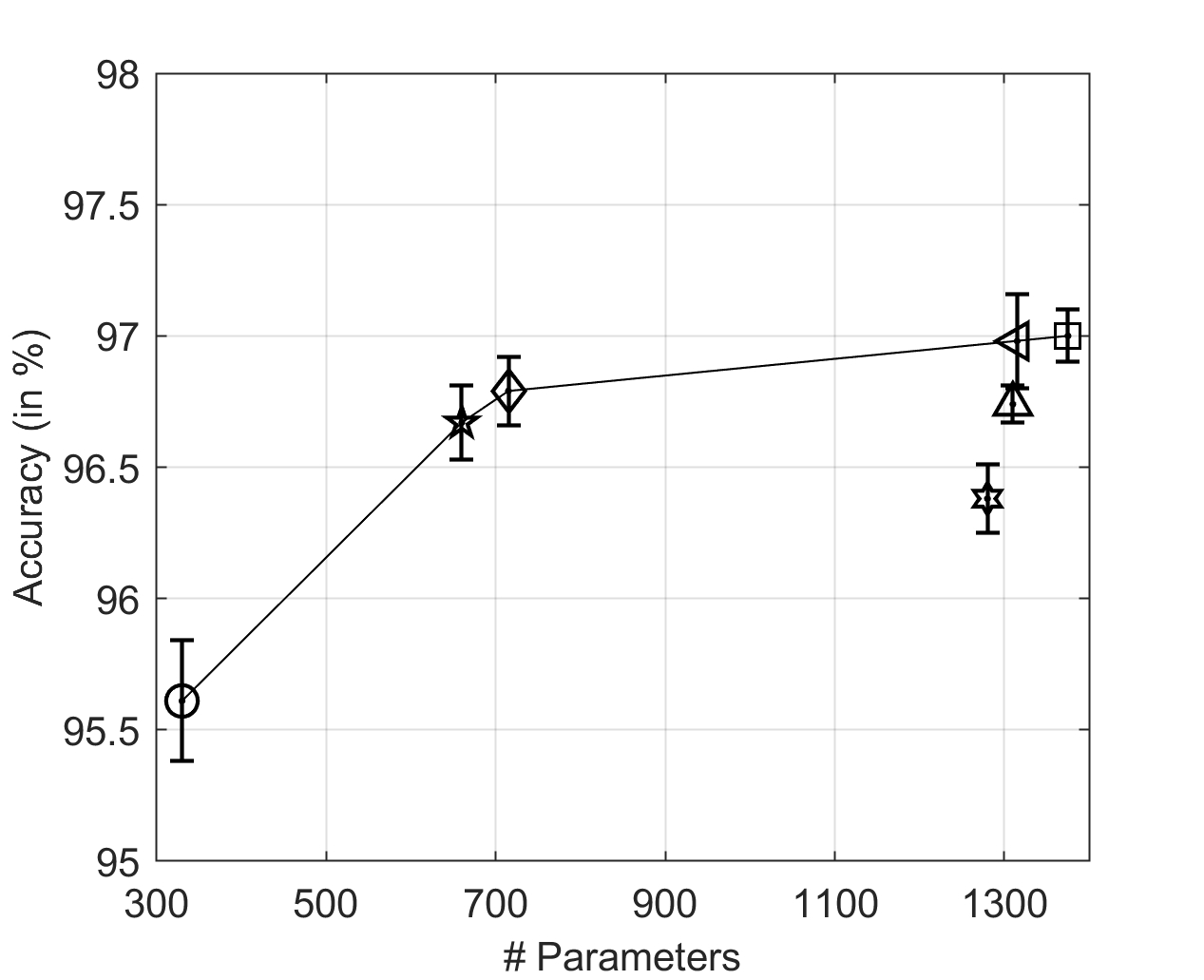}} &
\subfigure[Oriya]{\includegraphics[trim={0 0 1cm 0},clip,width=0.5\figwidth]{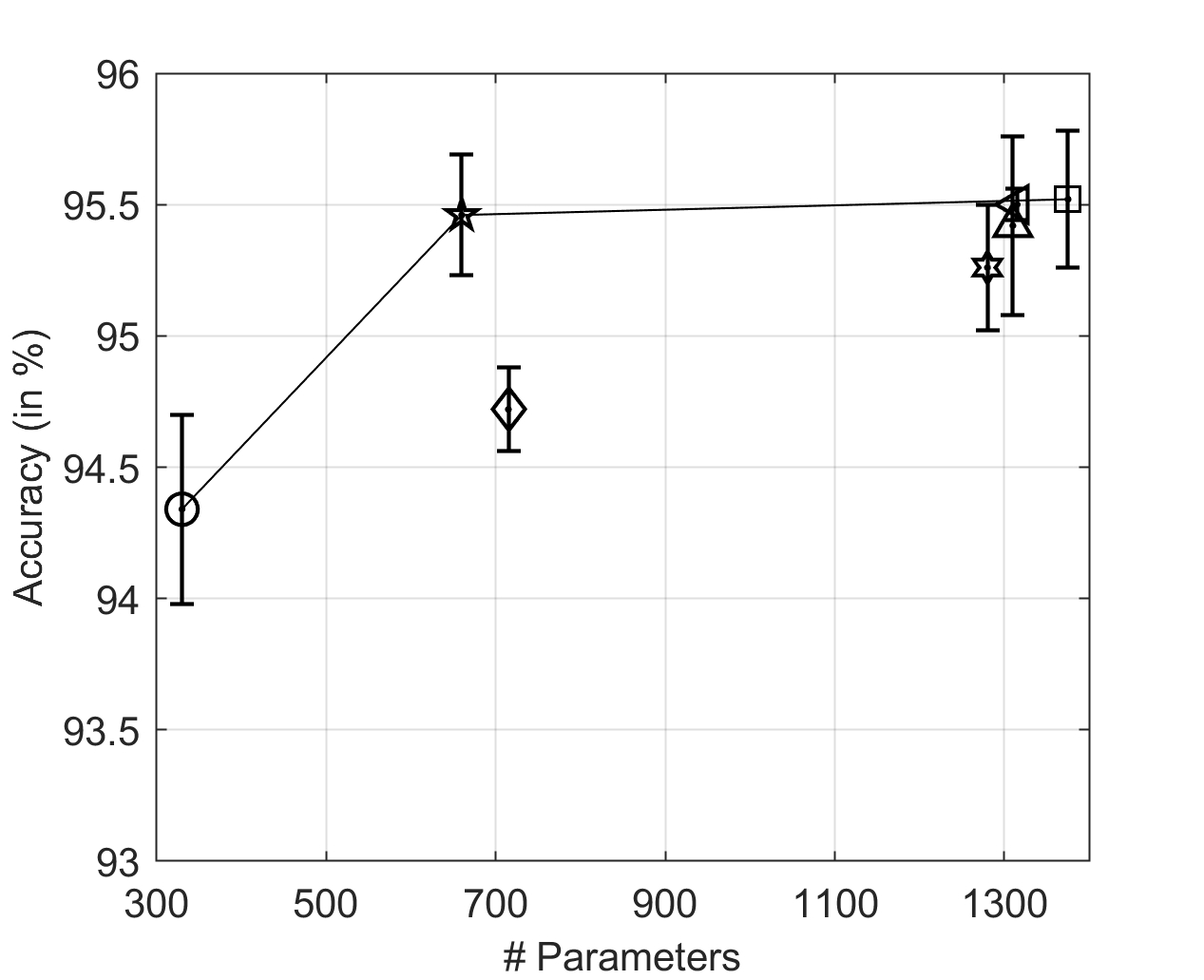}}
\end{tabular}
\caption{Accuracy (in \%) in relation to the number of parameters in convolutional layers.}
\label{fig:perf1}
\end{figure*}

\section{Discussion} 
\label{sec:discussion}

Three types of filters for convolutional neural networks have been proposed. These filters offer three key advantages: first, they reduce the total number of parameters to learn in the network; second they reduce the complexity of the forward operation, and third they provide linear phase filters which can have desirable properties for preserving the waveshape of the input signal. To validate the proposed approach, these filters have been tested on different databases of handwritten digits representing different types of challenges in relation to the number of examples for training, and the variability across examples for the chosen scripts. With the considered architecture, the results were aligned with the current state of the art, suggesting potential improvements with other more complex architectures.

Taking the results for testing accuracy for each of the four datasets and plotting them against the number of free parameters in each network, we can observe that the 
number of parameters does in general seem to be correlated with better accuracy. However, it is possible to note that for every dataset, a near quadrupling in the number parameters results in approximately a 1-2\% increase in accuracy. The figures that follow were developed using only the data from networks that actually learned. Looking at Fig.~\ref{fig:perf1}, for the MNIST dataset the biggest increase in accuracy occurs when moving from the \textsl{T1-T1} type filter to a \textsl{T2A-T2A} filter, and increase in parameters after that point has tiny gains. In fact, this seems to be true for all of the datasets in Fig.~\ref{fig:perf1}.

While the proposed approaches do not lead directly to the best accuracy, they provide key insights on the type of functions that need to be applied on images to extract robust descriptors. Typical image processing spatial filtering kernels embed filters with linear phase. In neural networks, the non-linearity is typically achieved through the activation function. We have shown that keeping a linear phase in the extracted filter slightly degrades the results while reducing substantially the number of weights in the network. Such an effect may be counter balanced by deep architectures. Furthermore, the difference in terms of number of examples per class from MNIST to the Oriya dataset and the pattern of performance across conditions suggest that the proposed filters have similar performance than the default condition when the number of examples is low. 

A linear phase filter will preserve the waveshape of the signal or component of the input signal. The proposed symmetrical filters can have implications in multiple applications that exploit transfer learning and in which it is necessary to provide linear phase filters. The waveshape is a relevant feature because a thresholding decision must be made on the waveshape in order to classify it. Therefore, preserving or recovering the originally transmitted waveshape is of critical importance, otherwise wrong thresholding decisions will be applied that would represent a bit error in a communications system. For instance, a CNN with linear phase convolutional layers can be used in phase-sensitive applications such as audio processing, radar signal processing, seismology~\cite{scherbaum1997}, where the waveshape of a returned radar signal may embed information about the target's properties. The filters that have been proposed can be employed in existing architectures to provide linear phase properties.

The current study analyzed the three types of filters separately for features extraction applied to the classification of handwritten digits. However, it is unknown how the combination of such filters would perform and what is the relationship between the type of filter and its place in the hierarchical architecture. Finally, this type of filter may be only used when there is a relationship between different input features, such as expressed through the notion of local neighborhood. For applications in which the convolution merges all the inputs from one dimension into multiple feature maps, i.e., when the size of one of the dimension of the filter has the same as one of the dimension of the input, then the proposed approach may not be considered.

\clearpage
\begin{figure*}[!]
\centering
\setlength{\tabcolsep}{0.1cm}
\renewcommand{\arraystretch}{2}
\begin{tabular}{@{}lccccccc@{}}	
& R-R & T1-T1 & T2A-T2A & T2B-T2B & T1-R & T2A-R & T2B-R \\
\rotatebox{90}{MNIST} &
\includegraphics[width=2.2cm]{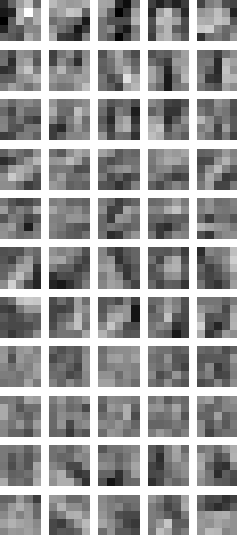} &
\includegraphics[width=2.2cm]{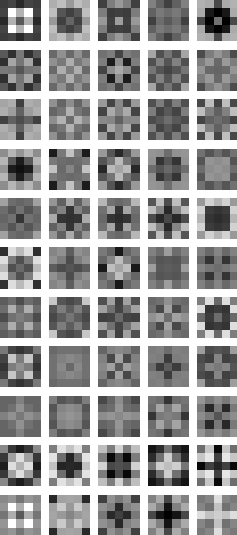} &
\includegraphics[width=2.2cm]{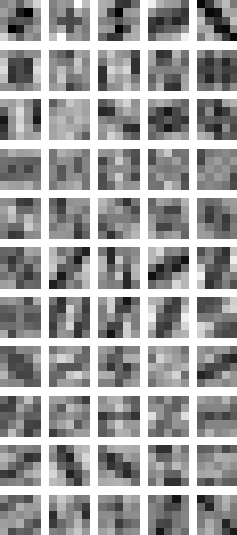} &
\includegraphics[width=2.2cm]{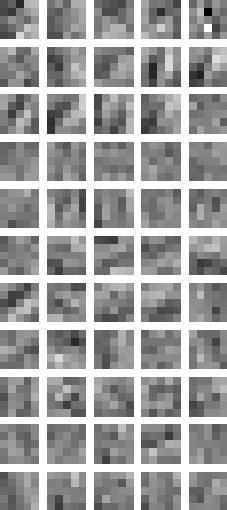} &
\includegraphics[width=2.2cm]{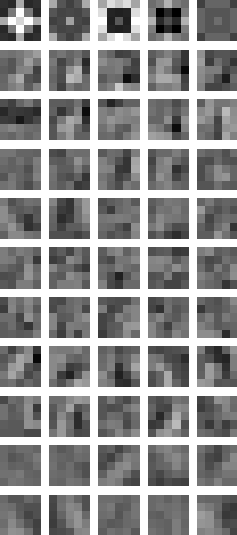} &
\includegraphics[width=2.2cm]{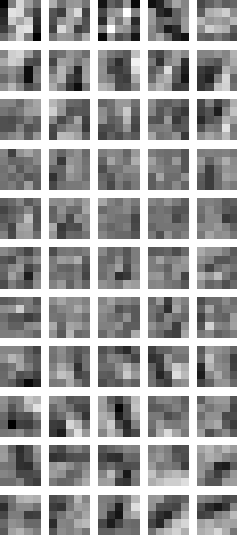} &
\includegraphics[width=2.2cm]{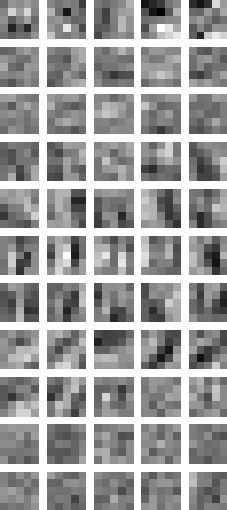} \\
\rotatebox{90}{Bangla} &
\includegraphics[width=2.2cm]{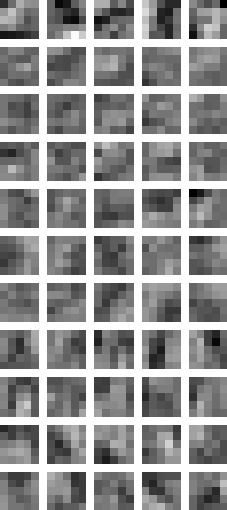} &
\includegraphics[width=2.2cm]{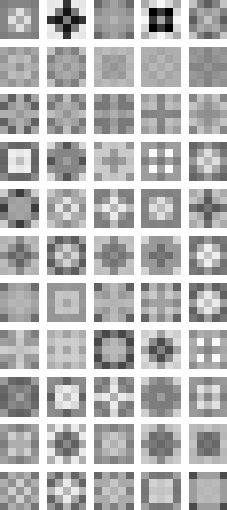} &
\includegraphics[width=2.2cm]{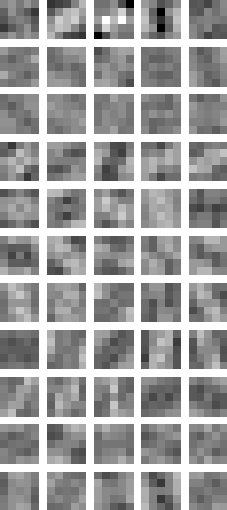} &
\includegraphics[width=2.2cm]{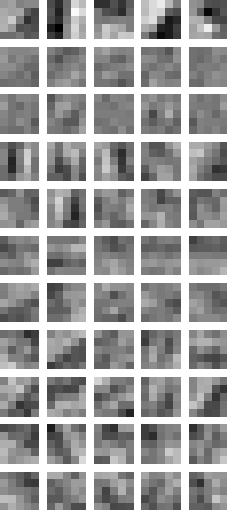} &
\includegraphics[width=2.2cm]{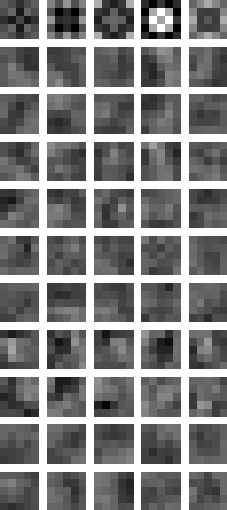} &
\includegraphics[width=2.2cm]{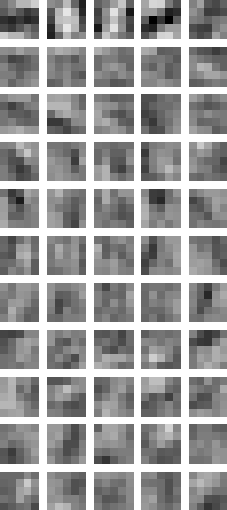} &
\includegraphics[width=2.2cm]{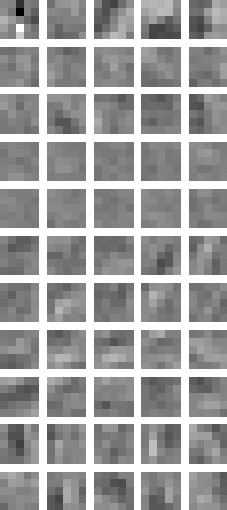} \\
\rotatebox{90}{Devanagari} &
\includegraphics[width=2.2cm]{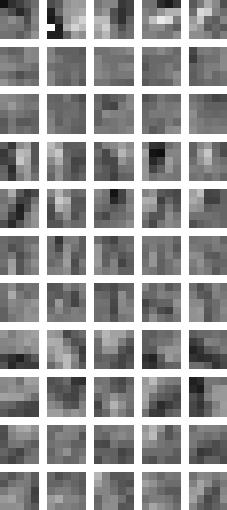} &
\includegraphics[width=2.2cm]{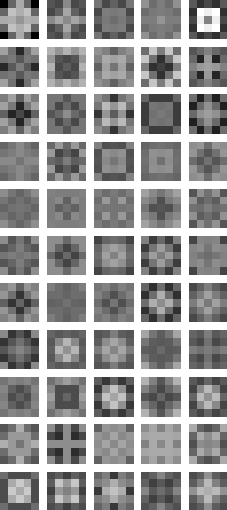} &
\includegraphics[width=2.2cm]{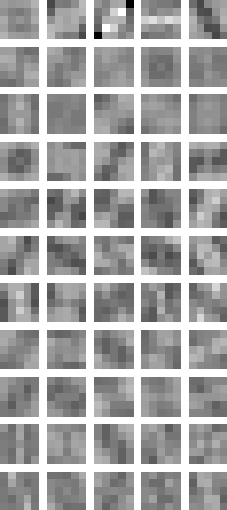} &
\includegraphics[width=2.2cm]{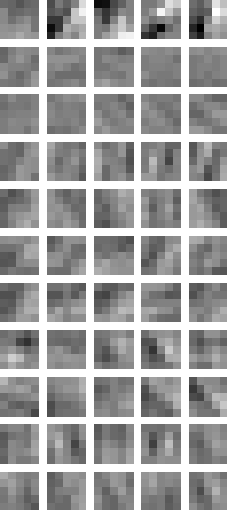} &
\includegraphics[width=2.2cm]{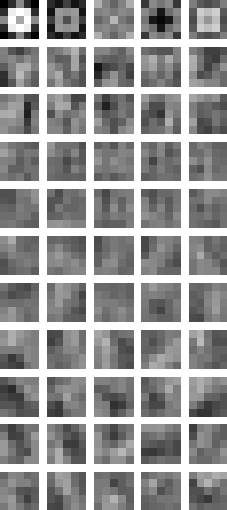} &
\includegraphics[width=2.2cm]{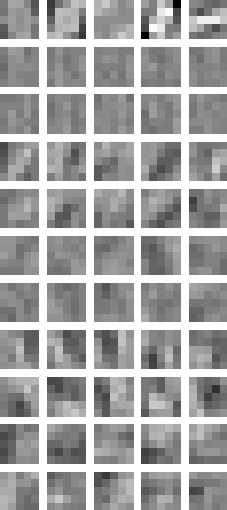} &
\includegraphics[width=2.2cm]{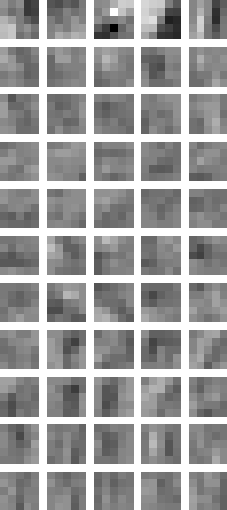} \\
\rotatebox{90}{Oriya} &
\includegraphics[width=2.2cm]{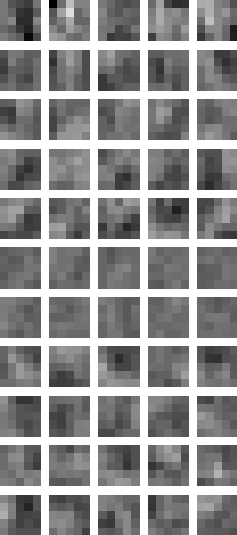} &
\includegraphics[width=2.2cm]{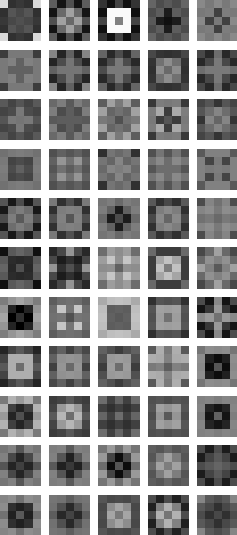} &
\includegraphics[width=2.2cm]{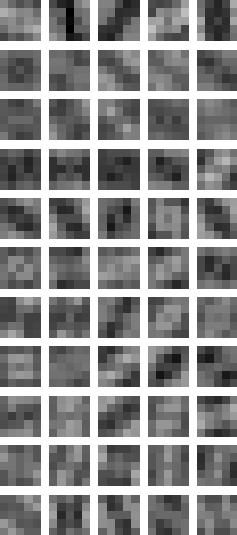} &
\includegraphics[width=2.2cm]{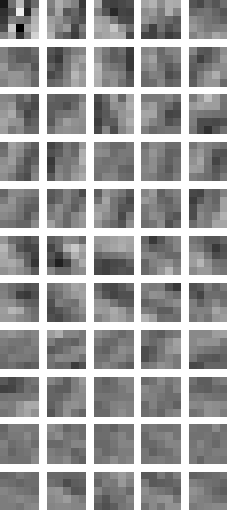} &
\includegraphics[width=2.2cm]{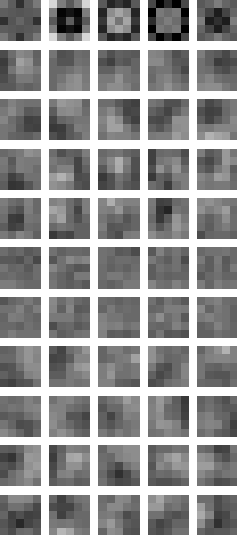} &
\includegraphics[width=2.2cm]{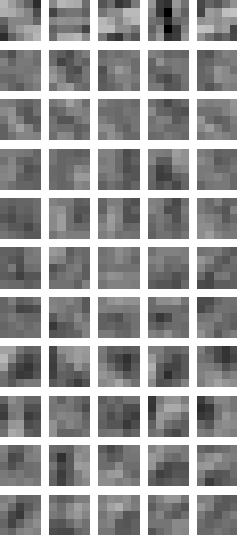} &
\includegraphics[width=2.2cm]{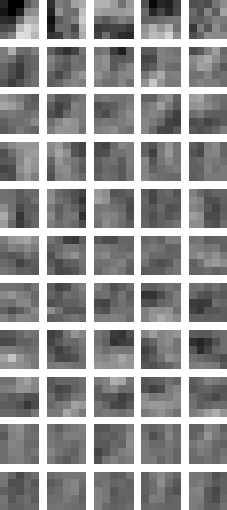} 
\end{tabular}
\caption{Learned filters for each condition and each datatset. In each block, the first row of five filters in each image corresponds to the filters learned in the first convolutional layer, the remaining rows correspond to the filters learned in the second convolutional layer. Each filter is individually scaled for a graylevel representation, as a function of its minimum and maximum values.}
\end{figure*}
\clearpage

\section{Conclusion} 
\label{sec:conclusion}

Deep learning approaches and especially convolutional neural networks have a high impact on society through their use in a large number of pattern recognition tasks. With their high performance, it is necessary to get some insights about their behavior and the advantages that they provide compared to more traditional approaches rooted in image and signal processing. A key challenge is to find the ideal frontier between what has to be learned and what can be determined analytically. In this paper, we have proposed a novel category of constraints for training convolutional layers that provide the linear phase property that can be found in typical FIR filters of Type I and III. Such an approach provides a substantial decrease of parameters, an increase in speed by reducing the complexity of the forward operation, and relatively equivalent performance compared to the traditional approach. Future works will be carried out to examine the behavior corresponding to the combinations of the such symmetrical filters.

\bibliographystyle{IEEEtran}
\bibliography{refs}

\end{document}